\documentclass[iicol,sn-mathphys-num]{sn-jnl}

\usepackage{graphicx}%
\usepackage{multirow}%
\usepackage{amsmath,amssymb,amsfonts}%
\usepackage{amsthm}%
\usepackage[title]{appendix}%
\usepackage{xcolor}%
\usepackage{textcomp}%
\usepackage{manyfoot}%
\usepackage{booktabs}%
\usepackage{subfigure}%
\usepackage{algorithm}%
\usepackage{algorithmic}
\usepackage{listings}%
\usepackage{url}
\usepackage{multicol}

\theoremstyle{thmstyleone}%
\newtheorem{theorem}{Theorem}
\theoremstyle{thmstyletwo}%

\theoremstyle{thmstylethree}%

\begin{document}

\title[Article Title]{
AI-enabled Satellite Edge Computing: A Single-Pixel Feature based Shallow Classification Model for Hyperspectral Imaging
}


\author[1,3]{\fnm{Li} \sur{Fang}}
\equalcont{These authors contributed equally to this work.}

\author[2,1]{\fnm{Tianyu} \sur{Li}}
\equalcont{These authors contributed equally to this work.}

\author[1,3]{\fnm{Yanghong} \sur{Lin}}

\author[1]{\fnm{Shudong} \sur{Zhou}}

\author*[1,3]{\fnm{Wei} \sur{Yao}}\email{wyao@iue.ac.cn}

\affil[1]{\orgdiv{State Key Laboratory of Regional and Urban Ecology}, \orgdiv{Institute of Urban Environment}, \orgname{Chinese Academy of Sciences}, \orgaddress{\city{Xiamen}, \country{China}}}

\affil[2]{\orgdiv{School of Resources and Environmental Sciences}, \orgname{Wuhan University}, \orgaddress{\city{Wuhan}, \country{China}}}

\affil[3]{\orgname{University of Chinese Academy of Sciences}, \orgaddress{\city{Beijing}, \country{China}}}


\abstract{
As the important component of the Earth observation system, hyperspectral imaging satellites provide high-fidelity and enriched information for the formulation of related policies due to the powerful spectral measurement capabilities. However, the transmission speed of the satellite downlink has become a major bottleneck in certain applications, such as disaster monitoring and emergency mapping, which demand a fast response ability. We propose an efficient AI-enabled Satellite Edge Computing paradigm for hyperspectral image classification, facilitating the satellites to attain autonomous decision-making. To accommodate the resource constraints of satellite platforms, the proposed method adopts a lightweight, non–deep learning framework integrated with a few-shot learning strategy. Moreover, onboard processing on satellites could be faced with sensor failure and scan pattern errors, which result in degraded image quality with bad/misaligned pixels and mixed noise. To address these challenges, we develop a novel two-stage pixel-wise label propagation scheme that utilizes only intrinsic spectral features at the single pixel level without the necessity to consider spatial structural information as requested by deep neural networks. In the first stage, initial pixel labels are obtained by propagating selected anchor labels through the constructed anchor–pixel affinity matrix. Subsequently, a top-$k$ pruned sparse graph is generated by directly computing pixel-level similarities. In the second stage, a closed-form solution derived from the sparse graph is employed to replace iterative computations. Furthermore, we developed a rank constraint-based graph clustering algorithm to determine the anchor labels.
}




\maketitle

\section{Introduction}
Earth observation missions are fundamental for long-term monitoring and research on global environmental change. To obtain reliable observation data, numerous satellites have been deployed in orbit, providing essential information to support policy formulation and decision-making. With the advent of high spatial and radiometric resolutions hyperspectral sensors, hyperspectral imaging missions have become increasingly critical due to their capability to capture continuous spectral information from the visible to the short-wave infrared range\cite{shi2025few, shi2025multil, shi2022explainable, zhang20191d, fang2025pcet}. 
The procurement of earth observation products from hyperspectral satellites involves a two-stage process: data transmission onboard the satellite and subsequent offline processing on terrestrial systems\cite{zhu2023collaborative}. This conventional ground-based interaction exhibits a delay in response due to the limit of transmission speed\cite{langer2023robust, jiang2023space}. Especially in certain critical emergency scenarios such as disaster monitoring\cite{arias2019hyperspectral} and emergency mapping\cite{kruse2014multispectral, he2024ub}, this procedure is prolonged and does not satisfy the demand for a rapid response. Thus, onboard processing of the acquired hyperspectral images in satellites remains a matter of significant importance and research interest.

The key to satellite onboard processing is to integrate artificial intelligence algorithms with satellite platform. Researchers have recently taken up onboard processing. Despite lacking a hyperspectral spectrometer, OPS-SAT serves as a pioneering validation platform to assess onboard processing capabilities\cite{evans2014ops}. To investigate the effects of modifying the quality of image data on deep learning models applied in Earth observation satellites, \cite{nalepa2021towards} conducted simulation experiments from the perspective of atmospheric conditions and noise interference. The $\Phi$-sat-1 mission \cite{giuffrida2021varphi} from the European Space Agency (ESA) illustrates the use of artificial intelligence in Earth observation. It is the first attempt to deploy a deep convolutional network on a satellite for the purpose of cloud segmentation. \cite{langer2023robust} focused on small satellites and assert that the critical factor in surmounting the limitations imposed by data bandwidth is the pre-processing of data onboard the satellite before transmission. Based on this idea, they proposed the \textit{Hyper-Spectral Small Satellite for Ocean Observation} (HYPSO) mission. In order to achieve real-time online classification of hyperspectral images, \cite{morcillo2024parametric} implemented the $k$-means algorithm as the onboard processing method. They also engineered specialized hardware to take advantage of various acceleration technologies aimed at minimizing overhead and optimizing performance. \cite{justo2024semantic} presents a lightweight deep learning model and substantiates its efficacy through validation by the HYPSO-1 and \textit{Earth Observing}-1 (EO-1) missions, whose main task is to segment the sea, land and cloud. Currently, onboard processing for hyperspectral images is focusing mainly on cloud segmentation and basic categorical classification, making it hard to apply to complex scenes.

The onboard processing of satellite hyperspectral images, in contrast to traditional post-downlink operations, is confronted with numerous challenges. First, compared to high-power and high-performance processors used in terrestrial settings, the computational resources available onboard satellites are significantly limited, especially for small satellites\cite{denby2019orbital}\cite{furano2020towards}\cite{george2018onboard}. This is mainly attributable to the inherent limitations imposed by the size, mass, cost, and power of the satellite. Recent advances in deep learning algorithms, especially large models, predominantly necessitate the utilization of high-performance computing platforms, thereby conflicting with the limited computational resources available on satellite systems\cite{zhang2024uav}. Consequently, the deployment of deep learning models onboard satellites is much more challenging and less desirable. The lack of large-scale remote sensing datasets also results in an inadequate foundation for deep learning models to formulate reasonable judgments in unanticipated situations\cite{audebert2019deep, sun2020deep}. Moreover, image quality degradation due to bad pixels\cite{kieffer1996detection}, mixed noise (e.g., Gaussian noise, poisson noise)\cite{zhang2019hybrid}\cite{aggarwal2016hyperspectral}, and misaligned pixels\cite{santos2022geometric} are common problems encountered in widely used spaceborne imaging spectrometers. Bad pixels are often denoted as impulse noise or salt-and-pepper noise ,which appear as random white or black pixels in images due to sensor operating failures or saturation of the scanning unit \cite{nalepa2021towards}\cite{zhang2019hybrid}\cite{zhuang2021fasthymix}. As the operational 
\begin{figure}[htbp]
	\centering
	\subfigure[]{
		\begin{minipage}[t]{0.3\linewidth}
			\centering
			\includegraphics[width=0.99\linewidth]{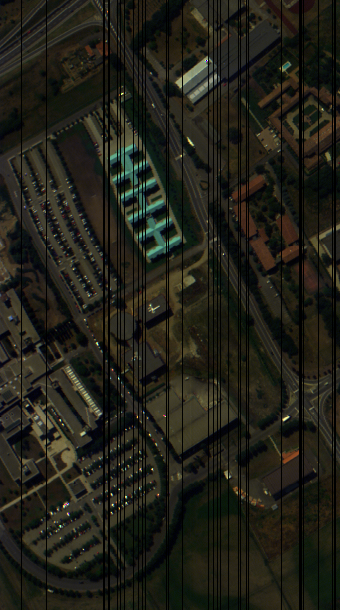}\\
   \end{minipage}}
   	\subfigure[]{
		\begin{minipage}[t]{0.3\linewidth}
			\centering
			\includegraphics[width=0.99\linewidth]{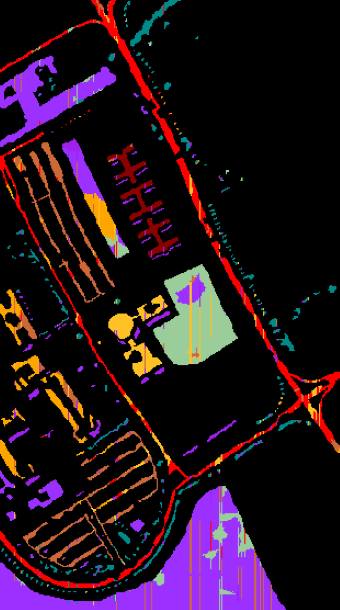}\\
			\end{minipage}}
   	\subfigure[]{
		\begin{minipage}[t]{0.3\linewidth}
			\centering
			\includegraphics[width=0.99\linewidth]{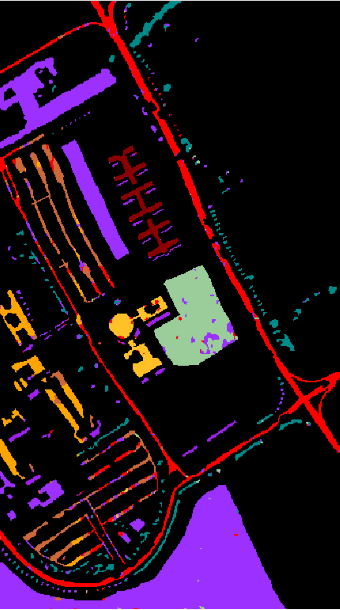}\\
			\end{minipage}}
	\centering
	\caption{A comparison of the classification performance of the proposed method and previous method under the interference of strip noise. (a) The falsecolor image of Pavia University with deadline noise. (b) The classification map with SGL\cite{sellars2020superpixel}. (c) The classification map using our method.}
	\label{fig:bad pixel}
\end{figure}
duration and frequency of the device activation cycles increase, these bad pixels will grow significantly. The occurrence of bad pixels and mixed noise corrupts the spectral bands and significantly degrades the performance of conventional classification methods based on machine learning, which typically exhibit limited robustness as well\cite{ma2024spatial}, as shown in Fig.\ref{fig:bad pixel}. Pixel misalignment is another typical phenomenon for satellite image acquisition. For example, the scan-line corrector (SLC) for the ETM+ sensor on board Landsat $7$, which compensates for the forward motion of the satellite, failed permanently \cite{chen2011simple, wang2021filling}. In the absence of a functional SLC, the acquired images exhibit wedge-shaped gaps that vary in width, ranging from a single pixel at the image nadir to approximately $12$ pixels towards the peripheries of the scene. Gap pixels could comprise approximately 22\% of all pixels in the image. With such misaligned scan lines and pixels, methods that rely on spatial neighborhood contexts to extract features, such as patch-based deep learning and conventional spatial-spectral methods, could obtain wrong spatial structural information, thus misleading the classification model. Hence, onboard satellite processing is essential to retain a high-efficiency lightweight model structure and maintain strong robustness to adverse effects of bad pixels, mixed noise, and pixel misalignment. 

To overcome the aforementioned challenges, this paper proposes a non deep learning-based hyperspectral image classification algorithm designed for deployment on low power consumption , low-computational-capacity satellite platforms. Specifically, we conceptualize a two-stage pixel-wise label propagation algorithm to disseminate the labels of selected anchors to unlabeled pixels. We are the first to propose only making use of single-pixel-level intrinsic spectral features to avoid being misled by poor / bad pixels in the local neighborhood and inaccurate spatial structural information during onboard processing. The label of the selected anchors by $k$-means is first propagated through the anchor graph to determine the initial pixel label associated with an affinity matrix, which is pruned to form a sparse graph with refined similarity between pixels. Then, during the second stage of label propagation, the initial pixel label is updated by a closed-form solution based on the derived sparse graph to achieve a more precise classification result. In addition, a rank-constraint-based graph clustering algorithm is developed to annotate the anchors instead of manual intervention. By reorganizing the anchor-anchor graph, the proposed algorithm shows a high adaptability to various complex scenes. In general, the proposed method for onboard hyperspectral image classification exhibits linear-time complexity and better robustness.

\section{Result}
\subsection{Benchmark}
To demonstrate the effectiveness of the proposed method, we conducted exhaustive experiments on three widely used hyperspectral image datasets.
\subsubsection{Indian Pines}
It represents one of the first hyperspectral image datasets for classification, captured by the Airborne Visual Infrared Imaging Spectrometer (AVIRIS) in 1992. The dataset has a size of $145\times145$ and $220$ spectral bands, of which 200 are retained after the exclusion of the $20$ water absorption bands. The dataset has a spatial resolution of approximately $20$ meters and includes $16$ land cover categories. Consequently, this increases the probability of generating mixed pixels, thus complicating the classification process.

\subsubsection{Salinas}
The dataset was acquired by the AVIRIS sensor within the Salinas Valley in California. It has a resolution of $512\times217$ and $224$ spectral bands. The original $224$ spectral bands included $20$ bands related to water absorption. After discarding these $20$ bands, the remaining $204$ bands were retained for classification. It has $16$ different land cover categories.

\subsubsection{Pavia University}
The dataset was obtained using the Reflective Optics System Imaging Spectrometer (ROSIS) sensor in the region of Pavia, located in northern Italy. The imagery comprises a spatial resolution of $610\times 340$ pixels and $103$ spectral bands. This dataset has been classified into $9$ classes, comprising a total of $42,776$ annotated samples.

\begin{table*}[htbp]
    \centering
    \caption{Per-Class Accuracies, AAs, OAs, and Kappa Coefficients of Different Methods on the Indian Pines Dataset with Five Training Examples of Each Class}
    \label{table:ip}
        \resizebox{\linewidth}{!}{ 
    \begin{tabular}{@{}ccccccccc@{}}
    \toprule
    class &CNN-PPF\cite{li2016hyperspectral} &GFHF\cite{10.5555/1104523} &PL\cite{calder2020poisson} &FSS\cite{zhong2020fusion} &SGL\cite{sellars2020superpixel} &STSE-DWLR\cite{zheng2019hyperspectral} &DSSPL\cite{zhong2021dynamic} &Ours \\ \midrule
    1 &0.6610 &0.8609 &0.9130 &0.7715 &0.9652 &0.9848 &\textbf{0.9913} &0.9778 \\
    2 &0.0806 &0.5564 &0.4863 &0.8388 &0.4856 &0.4791 &0.7078 &\textbf{0.9406} \\
    3 &0.0856 &0.5141 &0.4592 &0.7759 &0.5749 &0.6425 &0.7829 &\textbf{0.9042} \\
    4 &0.5746 &0.7460 &0.8468 &0.7965 &0.8384 &0.8350 &0.8540 &\textbf{0.9744} \\
    5 &0.0720 &0.7714 &0.7979 &0.6907 &0.8110 &0.8025 &0.8170 &\textbf{0.9561} \\
    6 &0.2265 &0.7011 &0.8192 &0.8138 &0.8301 &0.9116 &0.8940 &\textbf{1.0000} \\
    7 &0.4435 &0.9714 &0.9429 &0.7264 &0.9679 &0.9786 &0.9821 &\textbf{1.0000} \\
    8 &0.5476 &0.9920 &0.9973 &0.9129 &0.9969 &\textbf{0.9994} &0.9987 &0.9936 \\
    9 &0.5800 &0.9525 &0.8647 &0.6610 &\textbf{1.0000} &0.9900 &0.9900 &0.0000 \\
    10 &0.3065 &0.7213 &0.7267 &0.7508 &0.8110 &0.6989 &0.8148 &\textbf{1.0000} \\
    11 &0.2621 &0.6360 &0.6475 &0.9042 &0.6610 &0.8329 &0.8290 &\textbf{0.9496} \\
    12 &0.0954 &0.6440 &0.6125 &0.7921 &0.6857 &0.7764 &0.7248 &\textbf{0.9915} \\
    13 &0.6010 &0.8951 &0.9906 &0.8855 &0.9951 &0.9951 &\textbf{0.9951} &0.9901 \\
    14 &0.7606 &0.7324 &0.8283 &0.7483 &0.8603 &0.8304 &0.8712 &\textbf{1.0000} \\
    15 &0.4328 &0.7554 &0.7902 &0.6480 &0.8166 &0.8728 &0.8806 &\textbf{1.0000} \\
    16 &0.9000 &0.8903 &0.9892 &0.8963 &0.9806 &0.9514 &0.9903 &\textbf{1.0000} \\ \midrule
    AA &0.4144 &0.7057 &0.8033 &0.7883 &0.8300 &0.8576 &0.8702 &\textbf{0.9176} \\
    OA &0.3087 &0.6499 &0.7060 &0.8142 &0.7270 &0.8052 &0.8591 &\textbf{0.9663} \\
    Kappa &0.2488 &0.6641 &0.6592 &0.8028 &0.6940 &0.7868 &0.8332 &\textbf{0.9616} \\
    \bottomrule
    \end{tabular}}
\end{table*}

\subsection{Simulation of onboard processing under spaceborne conditions}
To test the effectiveness of the proposed two-stage label propagation for classification, we introduce CNN-PPF\cite{li2016hyperspectral}, GFHF\cite{10.5555/1104523}, PL\cite{calder2020poisson}, FSS\cite{zhong2020fusion}, SGL\cite{sellars2020superpixel}, STSE\_DWLR\cite{zheng2019hyperspectral}, and DSSPL\cite{zhong2021dynamic} as comparison methods. Among these methods, SGL and DSSPL employ graph-based semi-supervised classification techniques. It should be noted that DSSPL further incorporates the label propagation strategy. Furthermore, to evaluate the efficacy of the proposed clustering algorithms, we have introduced K-means, FCM, FSCAG\cite{wang2017fast}, SGCNR\cite{wang2019scalable}, HESSC\cite{rafiezadeh2020hierarchical}, NCSC\cite{cai2022superpixel}, SGLSC\cite{zhao2021superpixel}, $S^{3}$AGC\cite{chen2023spectral}, and SAPC\cite{jiang2024structured}. Within the aforementioned methods, SAPC is the most similar to the proposed approach, as it similarly utilizes anchor graph and rank constraint. To improve computational efficiency, we slice the input test set in a fixed interval $\theta$. $\theta$ was configured as $3000$, $4000$, and $4000$ for the Indian Pines, Salinas, and Pavia University datasets, respectively. The dimensions $d$ of the three datasets after PCA are set to $30$, $40$ and $50$, respectively. In experiments carried out within the semi-supervised framework, three evaluation metrics are introduced: average accuracy (AA), overall accuracy (OA) and the kappa coefficient. Following SAPC, in the clustering performance experiment, six evaluation metrics were employed: accuracy (ACC), kappa coefficient, normalized mutual information (NMI), purity, adjusted rand index (ARI) and F-score.

\subsubsection{Semi-supervised Classification Results}
Tables \ref{table:ip}, \ref{table:sa}, and \ref{table:pu} show the classification results of the proposed method in the Indian Pines, Salinas, and Pavia University dataset, where the anchor label is obtained from the ground truth. To ensure an equitable comparison of the performance of the proposed methods under the conditions of limited labeled data, five pixels are randomly sampled from each category to serve as the labeled data for the other seven compared methods. The proposed method demonstrates optimal classification performance on all datasets. In particular in the Indian Pines dataset, the proposed method achieves an increase in OA that exceeds 11\% compared to the state-of-the-art method. Even with the Pavia University dataset, which demonstrates the least favorable outcomes, the proposed method maintains comparable performance on par with the state-of-the-art method and exceeds it as the number of sampled pixels increases. Note that unlike existing methods that exhibit superior classification performance, our method exclusively utilizes the single-pixel-level spectral features, eschewing any incorporation of spatial information.

For the Indian Pines dataset, a remarkable improvement in classification accuracy is achieved compared to existing methods. The Indian Pines dataset exhibits a significant imbalance in class distribution, notably with the $7_{th}$ and $9_{th}$ classes comprising fewer than $30$ pixels. For most classes, the proposed method exhibits superior class accuracy compared to existing methods. However, the $9_{th}$ class is characterized by an extremely sparse pixel count, which leads to its exclusion during the anchor selection process, thus resulting in a classification accuracy of zero for this class. This phenomenon is effectively mitigated by increasing the number of anchors, for example, by sampling $10$ pixels per category. Meanwhile, this extreme imbalance among categories reduces the impact of this particular case of misclassification on OA. For the Salinas dataset, although existing methods already demonstrate high classification accuracy, the proposed method achieves a discernible degree of performance enhancement. Within the OA metric, the proposed method achieves an improvement of $2.1\%$. In contrast to the first two datasets, the Pavia University dataset has a smaller number of categories. Consequently, this led to a reduction in the number of anchors, which decreased from $80$ in the other datasets to $45$ in the Pavia University dataset. The integration of spatial information is particularly crucial in existing methods when sample features are limited, as it facilitates the attainment of high classification accuracy. However, our method still achieves results comparable to those of the state-of-the-art method in the absence of spatial information. Overall, the proposed method demonstrates excellent classification performance within a semi-supervised framework, independent of any reliance on spatial information.

\begin{table*}[htbp]
    \centering
    \caption{Per-Class Accuracies, AAs, OAs, and Kappa Coefficients of Different Methods on the SALINES Dataset with Five Training Examples of Each class}
    \label{table:sa}
        \resizebox{\linewidth}{!}{ 
    \begin{tabular}{@{}ccccccccc@{}}
    \toprule
    class &CNN-PPF\cite{li2016hyperspectral} &GFHF\cite{10.5555/1104523} &PL\cite{calder2020poisson} &FSS\cite{zhong2020fusion} &SGL\cite{sellars2020superpixel} &STSE-DWLR\cite{zheng2019hyperspectral} &DSSPL\cite{zhong2021dynamic} &Ours \\ \midrule
    1 &0.7510 &0.9985 &0.9985 &0.9370 &0.9905 &1.0000 &1.0000 &\textbf{1.0000} \\
    2 &0.5708 &0.9888 &0.9606 &0.9679 &0.9786 &1.0000 &\textbf{1.0000} &0.9952 \\
    3 &0.2613 &0.9958 &1.0000 &1.0000 &0.9613 &1.0000 &\textbf{1.0000} &0.9995 \\
    4 &0.1148 &\textbf{0.9753} &0.9799 &0.9625 &0.9799 &0.8911 &0.9714 &0.9483 \\
    5 &0.3869 &0.9585 &0.9599 &0.9814 &0.9677 &0.9942 &0.9564 &\textbf{0.9996} \\
    6 &0.7805 &0.9942 &0.9942 &0.9655 &0.9948 &0.9887 &\textbf{0.9981} &0.9977 \\
    7 &0.3983 &0.9971 &0.9961 &0.9439 &0.9960 &0.8517 &0.9884 &\textbf{0.9986} \\
    8 &0.2916 &0.9048 &0.8592 &0.9093 &0.8836 &0.9517 &0.9586 &\textbf{0.9917} \\
    9 &0.7770 &0.9991 &0.9495 &0.9162 &0.9802 &1.0000 &0.9985 &\textbf{1.0000} \\
    10 &0.1750 &0.7322 &0.7729 &0.9439 &0.7939 &0.8883 &0.9546 &\textbf{0.9655} \\
    11 &0.2826 &0.9562 &0.9522 &0.9516 &0.9355 &0.9326 &0.9583 &\textbf{0.9991} \\
    12 &0.2534 &1.0000 &0.9485 &1.0000 &1.0000 &1.0000 &\textbf{1.0000} &0.9964 \\
    13 &0.6232 &0.9771 &0.9796 &0.9780 &0.9781 &0.9152 &0.9028 &\textbf{0.9989} \\
    14 &0.5114 &0.9407 &0.9062 &0.9488 &0.9149 &0.9600 &0.9621 &\textbf{0.9953} \\
    15 &0.0334 &0.9266 &0.9603 &0.9313 &0.9670 &0.9783 &\textbf{0.9797} &0.9612 \\
    16 &0.3582 &0.9792 &0.9956 &0.9514 &0.9904 &1.0000 &\textbf{1.0000} &0.9922 \\
    \midrule
    AA &0.4106 &0.9590 &0.9508 &0.9556 &0.9570 &0.9595 &0.9647 &\textbf{0.9899} \\
    OA &0.3943 &0.9473 &0.9347 &0.9549 &0.9469 &0.9630 &0.9675 &\textbf{0.9885} \\
    Kappa &0.3499 &0.9414 &0.9276 &0.9498 &0.9411 &0.9589 &0.9638 &\textbf{0.9872} \\
    \bottomrule
    \end{tabular}}
\end{table*}

\subsubsection{Impact of the Number of Labeled Samples}
To investigate the influence of the varying number of labeled samples on the method, we established four sampling strategies comprising $3$, $5$, $10$, and $30$ samples per class. Table \ref{table:oa} presents the comparative performance of the proposed methodology and the existing approaches with respect to overall accuracy (OA) in four distinct sampling strategies. Given that the Indian Pines dataset contains fewer than $30$ pixels for the classes $7$ and $9$, a sampling size of $15$ pixels is used for these classes instead of the standard $30$ pixels. The analysis of the results reveals that an increase in the number of labeled samples substantively enhances the classification accuracy across all methods. It is significant to observe that the proposed method yields superior performance compared to all other methods in the four sampling numbers on the Indian Pines and Salinas datasets. Specifically, when using a sampling strategy of $30$ labeled pixels per class, our method achieves an overall accuracy of $99\%$. In the case of the Pavia University dataset, the proposed method exhibits a lower performance relative to the state-of-the-art method when employing sampling strategies of $3$ pixels per class. However, as the number of samples is increased, the accuracy of our method begins to improve significantly and eventually exceeds the performance of the state-of-the-art method. This phenomenon can be attributed to the fact that with an increased number of labeled pixels, the richness of the available spectral information is enhanced. In this scenario, despite the lack of spatial information, our method is able to take full advantage of the enhanced spectral features to achieve superior classification accuracy. By setting a different number of samples, it is demonstrated that our method exhibits superior capability in fully leveraging the augmented pixel compared to other methods.

\begin{figure*}[htbp]
	\centering
	\subfigure[]{
		\begin{minipage}[t]{0.18\linewidth}
			\centering
			\includegraphics[width=0.99\linewidth]{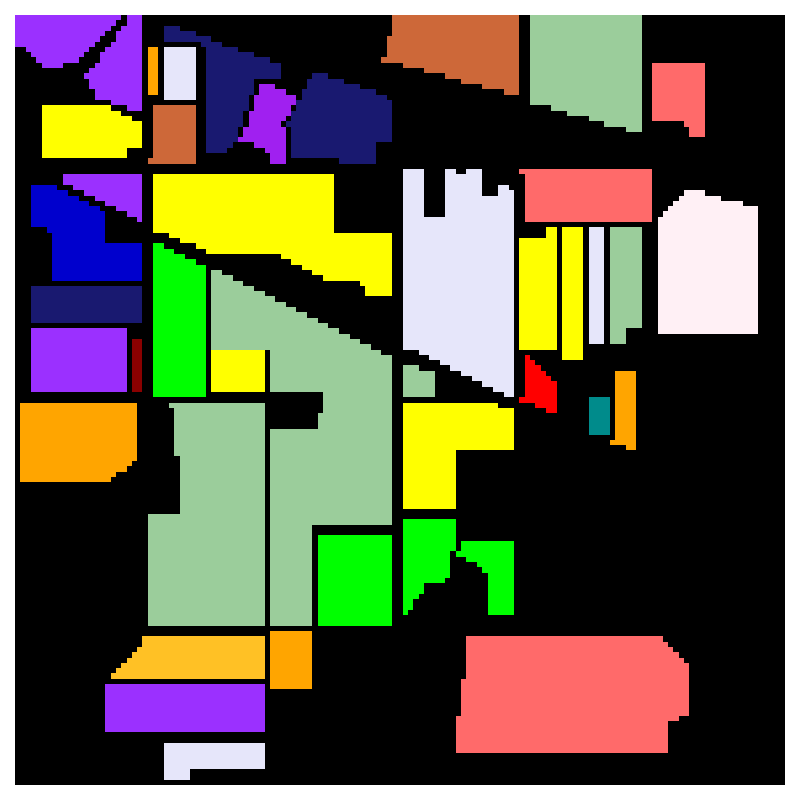}\\
   \end{minipage}}
   	\subfigure[]{
		\begin{minipage}[t]{0.18\linewidth}
			\centering
			\includegraphics[width=0.99\linewidth]{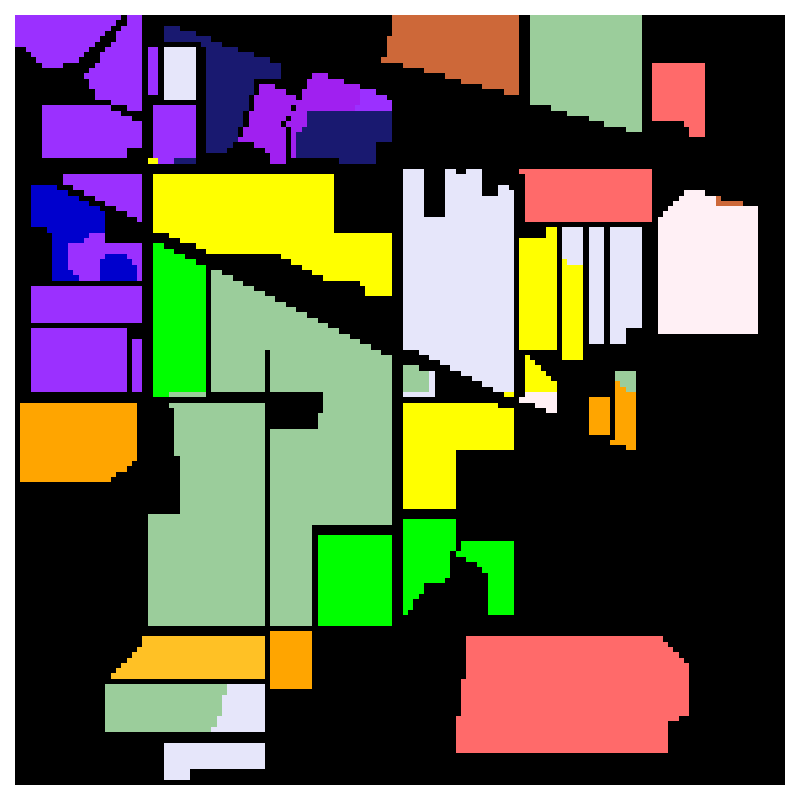}\\
			\end{minipage}}
   	\subfigure[]{
		\begin{minipage}[t]{0.18\linewidth}
			\centering
			\includegraphics[width=0.99\linewidth]{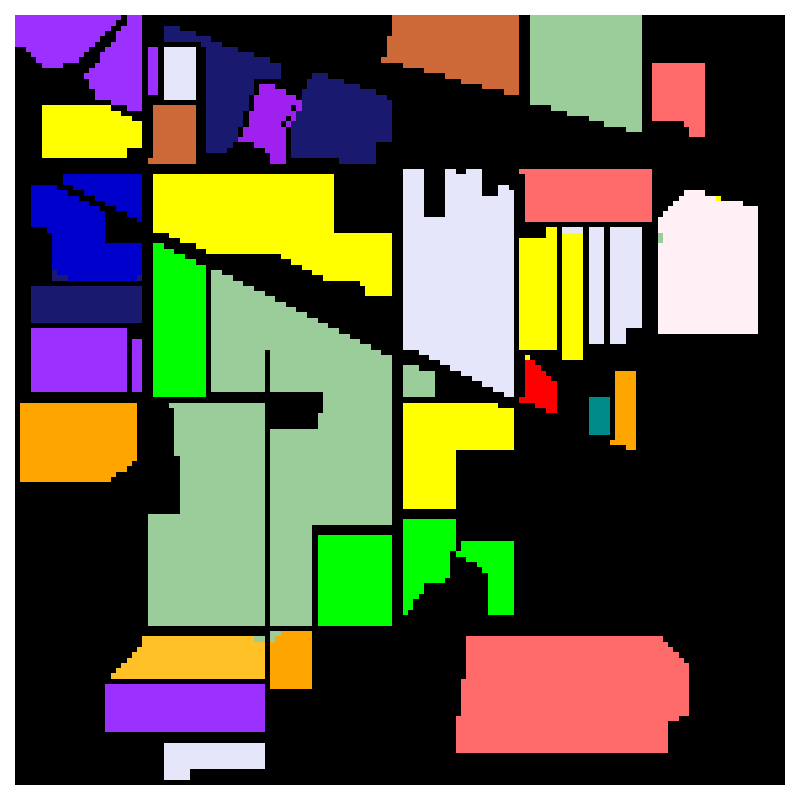}\\
			\end{minipage}}
               	\subfigure[]{
		\begin{minipage}[t]{0.18\linewidth}
			\centering
			\includegraphics[width=0.99\linewidth]{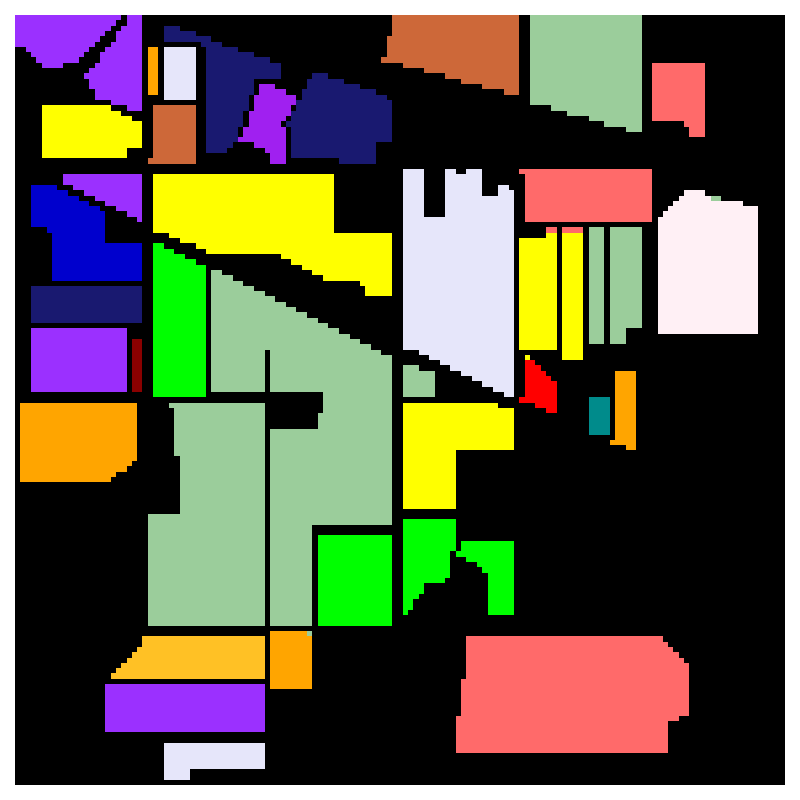}\\
			\end{minipage}}
               	\subfigure[]{
		\begin{minipage}[t]{0.18\linewidth}
			\centering
			\includegraphics[width=0.99\linewidth]{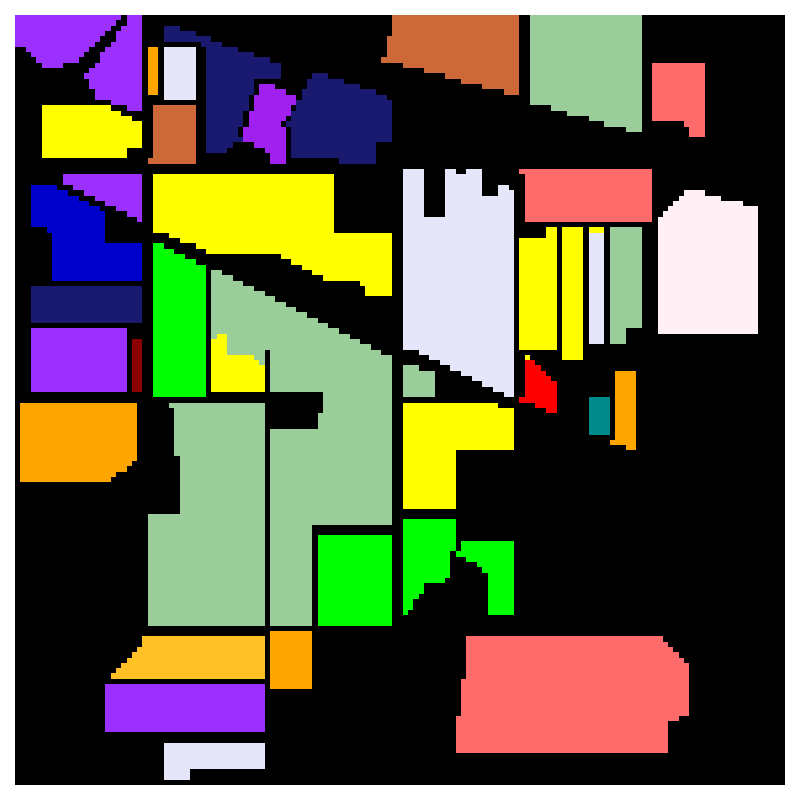}\\
			\end{minipage}}
	\centering
	\caption{Ground-truth map and classification maps on the Indian Pines dataset using different training samples per class. (a) ground truth. (b) $3$ training samples per class. (c) $5$ training samples per class. (d) $10$ training samples per class. (e) $30$ training samples per class.}
	\label{fig:ip}
\end{figure*}

\begin{figure*}[htbp]
	\centering
	\subfigure[]{
		\begin{minipage}[t]{0.18\linewidth}
			\centering
			\includegraphics[width=0.99\linewidth]{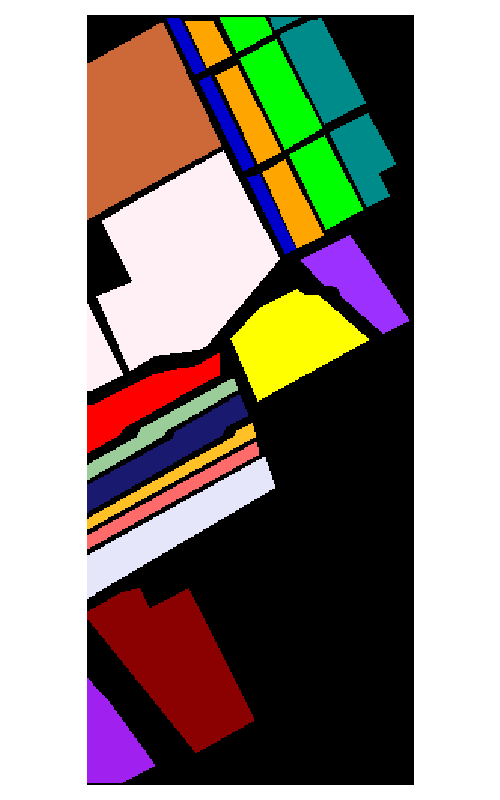}\\
   \end{minipage}}
   	\subfigure[]{
		\begin{minipage}[t]{0.18\linewidth}
			\centering
			\includegraphics[width=0.99\linewidth]{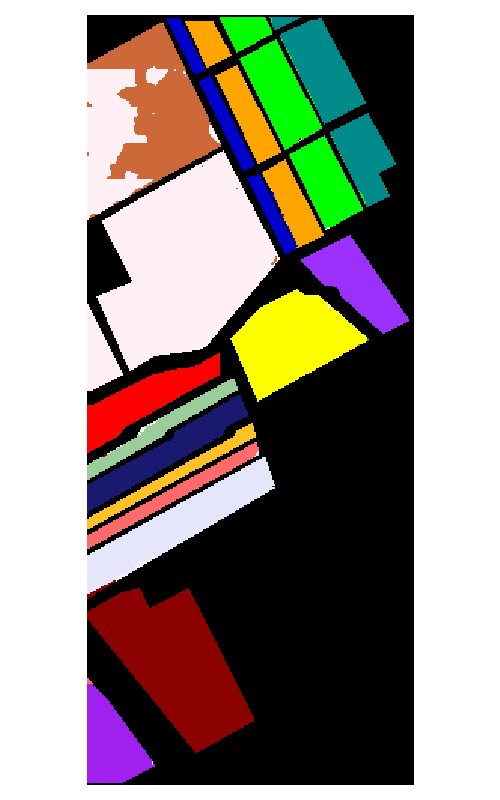}\\
			\end{minipage}}
   	\subfigure[]{
		\begin{minipage}[t]{0.18\linewidth}
			\centering
			\includegraphics[width=0.99\linewidth]{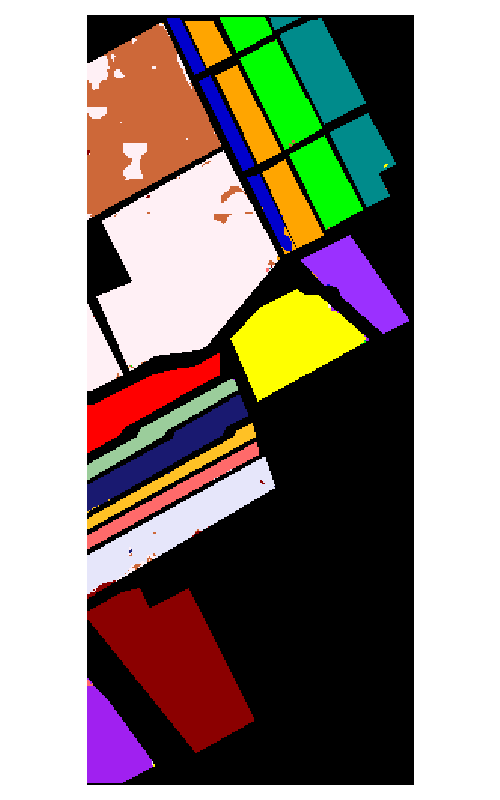}\\
			\end{minipage}}
               	\subfigure[]{
		\begin{minipage}[t]{0.18\linewidth}
			\centering
			\includegraphics[width=0.99\linewidth]{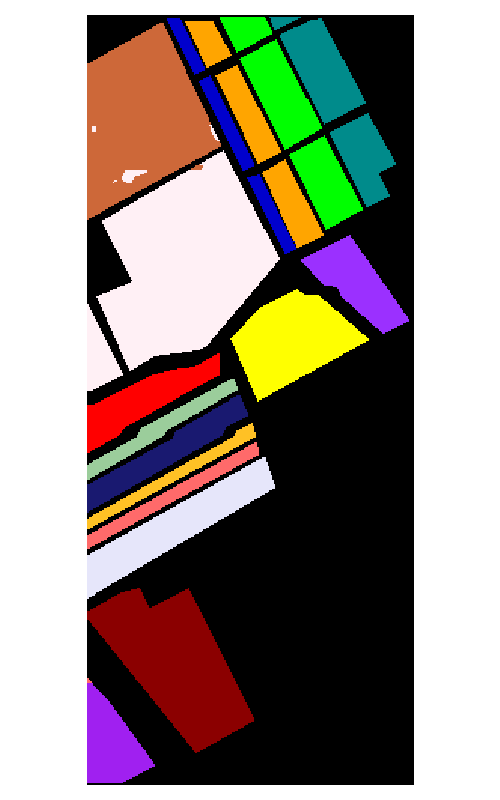}\\
			\end{minipage}}
               	\subfigure[]{
		\begin{minipage}[t]{0.18\linewidth}
			\centering
			\includegraphics[width=0.99\linewidth]{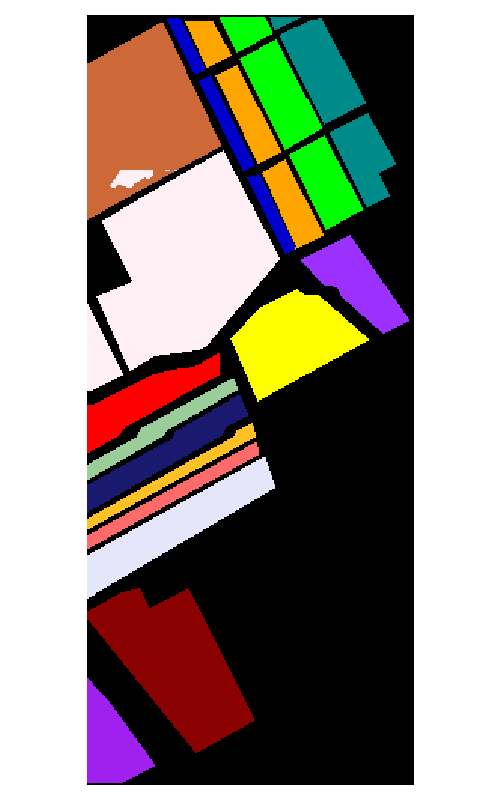}\\
			\end{minipage}}
	\centering
	\caption{Ground-truth map and classification maps on the Salinas dataset using different training samples per class. (a) ground truth. (b) $3$ training samples per class. (c) $5$ training samples per class. (d) $10$ training samples per class. (e) $30$ training samples per class.}
	\label{fig:sa}
\end{figure*}

\begin{figure*}[htbp]
	\centering
	\subfigure[]{
		\begin{minipage}[t]{0.18\linewidth}
			\centering
			\includegraphics[width=0.99\linewidth]{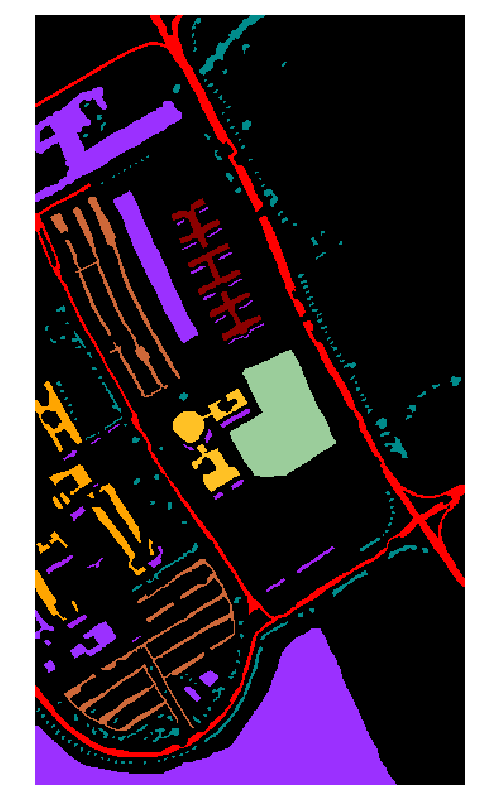}\\
   \end{minipage}}
   	\subfigure[]{
		\begin{minipage}[t]{0.18\linewidth}
			\centering
			\includegraphics[width=0.99\linewidth]{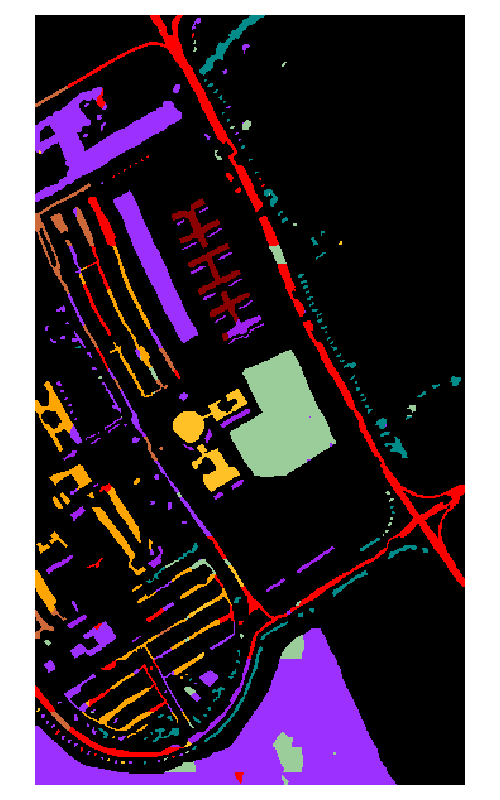}\\
			\end{minipage}}
   	\subfigure[]{
		\begin{minipage}[t]{0.18\linewidth}
			\centering
			\includegraphics[width=0.99\linewidth]{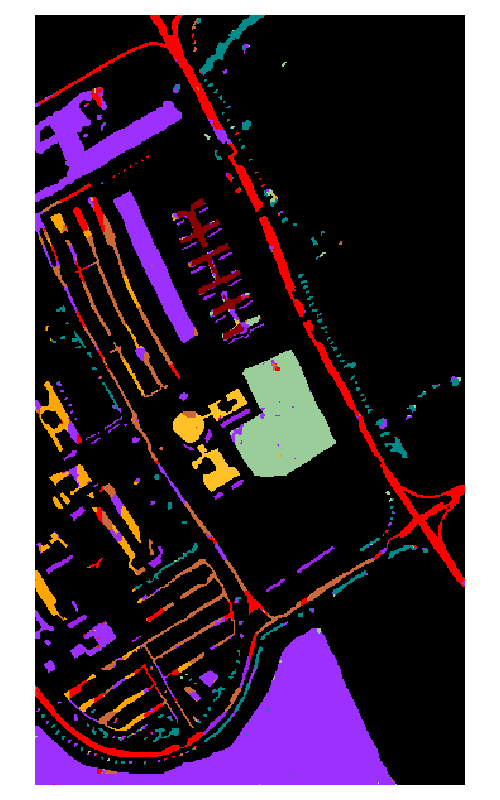}\\
			\end{minipage}}
               	\subfigure[]{
		\begin{minipage}[t]{0.18\linewidth}
			\centering
			\includegraphics[width=0.99\linewidth]{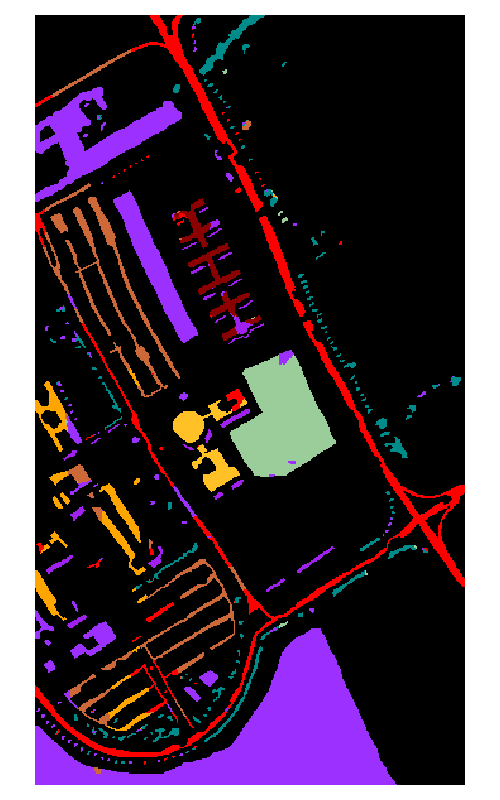}\\
			\end{minipage}}
               	\subfigure[]{
		\begin{minipage}[t]{0.18\linewidth}
			\centering
			\includegraphics[width=0.99\linewidth]{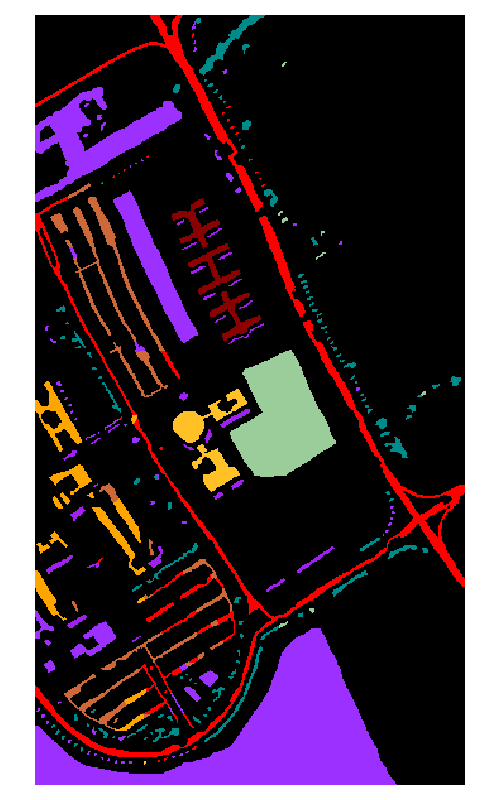}\\
			\end{minipage}}
	\centering
	\caption{Ground-truth map and classification maps on the Pavia University dataset using different training samples per class. (a) ground truth. (b) $3$ training samples per class. (c) $5$ training samples per class. (d) $10$ training samples per class. (e) $30$ training samples per class.}
	\label{fig:pu}
\end{figure*}

Furthermore, we present maps that illustrate the classification results of our method in varying numbers of training samples, as depicted in Fig.\ref{fig:ip}, Fig.\ref{fig:sa}, Fig.\ref{fig:pu}. It is evident that with an increment in the number of training samples, there is an improvement in classification accuracy, which is consistent with the findings of the quantitative analysis.

\subsubsection{Unsupervised Classification Results}
Table \ref{table:cluster} presents the results of the proposed clustering algorithm performed on the Salinas and Pavia University datasets. Our method achieves notable clustering efficacy in the Salinas dataset, displaying distinct advantages by utilizing a reduced number of anchors. Specifically, our method improves the metrics ACC, Kappa, NMI, Purity, ARI, and F-Score by $0.0988$, $0.1432$, $0.0379$, $0.1064$, $0.1393$ and $0.1897$, respectively. This demonstrates the effectiveness of the proposed clustering algorithm. However, in contrast to the Salinas dataset, the Pavia University dataset presents a more limited number of classes and exhibits a more acute issue of class imbalance. The existing state-of-the-art method, SAPC, demonstrates certain advantages through the integration of spatial information. It is evident that our method is affected by category imbalance when spatial features are not incorporated. The clustering algorithm shows a propensity towards larger categories while neglecting smaller ones, thereby leading to reduced Kappa and NMI values and an increase in Purity. Nevertheless, with respect to overall accuracy and the F-Score metric, the proposed method exhibits equivalent performance to the state-of-the-art method.

\begin{table*}[htbp]
    \centering
    \caption{Per-Class Accuracies, AAs, OAs, and Kappa Coefficients of Different Methods on the PAVIA UNIVERSITY Dataset with Five Training Examples of Each Class}
    \label{table:pu}
        \resizebox{\linewidth}{!}{ 
    \begin{tabular}{@{}ccccccccc@{}}
    \toprule
    class &CNN-PPF\cite{li2016hyperspectral} &GFHF\cite{10.5555/1104523} &PL\cite{calder2020poisson} &FSS\cite{zhong2020fusion} &SGL\cite{sellars2020superpixel} &STSE-DWLR\cite{zheng2019hyperspectral} &DSSPL\cite{zhong2021dynamic} &Ours \\ \midrule
    1 &0.7414 &0.7078 &0.7450 &0.7000 &0.8013 &0.7769 &\textbf{0.8072} &0.7399 \\
    2 &0.4258 &0.7907 &0.7905 &0.7947 &0.8080 &0.8225 &0.9273 &\textbf{0.9850} \\
    3 &0.7805 &0.9795 &0.9123 &0.9506 &\textbf{0.9986} &0.9812 &0.9524 &0.7624 \\
    4 &0.9143 &0.5192 &0.5348 &0.3285 &0.6478 &0.5944 &\textbf{0.8283} &0.7028 \\
    5 &0.9937 &0.8871 &0.8695 &0.9547 &0.9680 &\textbf{0.9854} &0.9631 &0.7640 \\
    6 &0.4460 &0.9676 &0.9337 &0.9738 &0.7775 &0.9774 &0.9929 &\textbf{0.9952} \\
    7 &0.9362 &0.8754 &0.8759 &0.9967 &\textbf{1.0000} &0.8842 &0.8806 &0.9503 \\
    8 &0.7063 &0.8463 &0.8798 &\textbf{0.8772} &0.7773 &0.7512 &0.8329 &0.5591 \\
    9 &0.9843 &0.9860 &0.9989 &0.3041 &0.9989 &0.9989 &\textbf{0.9989} &0.9715 \\
    \midrule
    AA &0.7698 &0.7744 &0.8078 &0.8378 &0.8653 &0.8769 &\textbf{0.8937} &0.8256 \\
    OA &0.5996 &0.7252 &0.7151 &0.8039 &0.8569 &0.8302 &0.8674 &\textbf{0.8721} \\
    Kappa &0.5213 &0.7774 &0.7641 &0.7774 &0.7608 &0.7884 &\textbf{0.8405} &0.8281 \\
    \bottomrule
    \end{tabular}}
\end{table*}

\begin{table*}[htbp]
    \centering
    \caption{OAs of Different Methods on the Indian Pines, Salinas, and Pavia University Datasets With Different Numbers of Training Pixels of Each Class}
    \label{table:oa}
        \resizebox{\linewidth}{!}{ 
    \begin{tabular}{@{}cccccccccc@{}}
    \toprule
      &Number  &CNN-PPF\cite{li2016hyperspectral} &GFHF\cite{10.5555/1104523} &PL\cite{calder2020poisson} &FSS\cite{zhong2020fusion} &SGL\cite{sellars2020superpixel} &STSE-DWLR\cite{zheng2019hyperspectral} &DSSPL\cite{zhong2021dynamic} &Ours \\ \midrule
    \multirow{4}{*}{Indian Pines} &3 &0.2169 &0.6056 &0.6251 &0.7043 &0.6882 &0.7514 &0.7924 &\textbf{0.8798} \\
     &5 &0.3087 &0.6499 &0.7060 &0.8142 &0.7270 &0.8052 &0.8591 &\textbf{0.9663} \\
     &10 &0.6132 &0.7125 &0.7238 &0.8691 &0.7695 &0.8829 &0.8810 &\textbf{0.9845} \\
     &30/15 &0.7986 &0.8026 &0.7948 &0.9666 &0.8570 &0.9625 &0.9433 &\textbf{0.9977} \\
    \midrule
     \multirow{4}{*}{Salinas} &3 &0.1168 &0.8124 &0.8347 &0.8528 &0.8831 &0.9259 &0.9358 &\textbf{0.9631} \\
     &5 &0.3943 &0.9473 &0.9347 &0.9549 &0.9469 &0.9630 &0.9675 &\textbf{0.9885} \\
     &10 &0.7765 &0.9547 &0.9412 &0.9789 &0.9617 &0.9745 &0.9799 &\textbf{0.9971} \\
     &30 &0.9051 &0.9723 &0.9622 &0.9860 &0.9695 &0.9814 &0.9887 &\textbf{0.9957} \\
    \midrule
     \multirow{4}{*}{Pavia University} &3 &0.4780 &0.7015 &0.7098 &0.7482 &0.7581 &0.7931 &\textbf{0.8419} &0.8242 \\
     &5 &0.5996 &0.7252 &0.7151 &0.8039 &0.8569 &0.8302 &0.8674 &\textbf{0.8721} \\
     &10 &0.7417 &0.7524 &0.7418 &0.8517 &0.8826 &0.9218 &0.9242 &\textbf{0.9292} \\
     &30 &0.8919 &0.8269 &0.8325 &0.9564 &0.9234 &0.9369 &0.9429 &\textbf{0.9514} \\
    \bottomrule
    \end{tabular}}
\end{table*}

\begin{table*}[htbp]
    \centering
    \caption{Clustering Performance of Different Methods on the Salinas, and Pavia University Datasets}
    \label{table:cluster}
        \resizebox{\linewidth}{!}{ 
    \begin{tabular}{@{}cccccccccccc@{}}
    \toprule
     Datasets &Metrics &K-means &FCM &FSCAG\cite{wang2017fast} &SGCNR\cite{wang2019scalable} &HESSC\cite{rafiezadeh2020hierarchical} &NCSC\cite{cai2022superpixel} &SGLSC\cite{zhao2021superpixel} &$S^{3}$AGC\cite{chen2023spectral} &SAPC\cite{jiang2024structured} &Ours \\ \midrule
    \multirow{6}{*}{Salinas} 
    &ACC $\uparrow$ &0.6456 &0.6065 &0.6590 &0.6224 &0.5293 &0.7497 &0.7664 &0.7465 &0.7881 &\textbf{0.8869}\\
    &Kappa $\uparrow$ &0.6044 &0.5671 &0.6212 &0.5790 &0.4787 &0.7206 &0.7407 &0.7181 &0.7637 &\textbf{0.9069} \\
    &NMI $\uparrow$ &0.7275 &0.7072 &0.7412 &0.7124 &0.6238 &0.8176 &0.8388 &0.7836 &0.8464 &\textbf{0.8843} \\
    &Purity $\uparrow$ &0.6640 &0.6686 &0.7070 &0.6688 &0.5881 &0.7638 &0.7679 &0.7521 &0.8056 &\textbf{0.9120} \\
    &ARI $\uparrow$ &0.5392 &0.4761 &0.5444 &0.4978 &0.4150 &0.6650 &0.6553 &0.6095 &0.6929 &\textbf{0.8322} \\
    &F-Score $\uparrow$ &0.5549 &0.5911 &0.5818 &0.5462 &0.4375 &0.5821 &0.6072 &0.6668 &0.6972 &\textbf{0.8869} \\
    \midrule
    \multirow{6}{*}{Pavia University} 
    &ACC $\uparrow$ &0.5370 &0.5263 &0.5288 &0.5019 &0.4790 &0.4392 &0.6245 &0.4855 &\textbf{0.6394} &0.6188\\
    &Kappa $\uparrow$ &0.4357 &0.4245 &0.4291 &0.4039 &0.3663 &0.2952 &0.4885 &0.3774 &\textbf{0.5294} &0.4433 \\
    &NMI $\uparrow$ &0.5321 &0.5291 &0.5588 &0.5068 &0.4982 &0.4355 &0.5410 &0.4767 &\textbf{0.6189} &0.4392 \\
    &Purity $\uparrow$ &0.6957 &0.6936 &0.7055 &0.6628 &0.6397 &0.6037 &0.6769 &0.6341 &0.7132 &\textbf{0.8916} \\
    &ARI $\uparrow$ &0.3255 &0.3167 &0.3306 &0.2982 &0.3020 &0.2905 &0.5063 &0.2616 &\textbf{0.4491} &0.4128 \\
    &F-Score $\uparrow$ &0.5079 &0.5036 &0.5309 &0.4767 &0.4130 &0.3656 &0.4201 &0.4815 &0.5602 &\textbf{0.6108} \\ 
    \bottomrule
    \end{tabular}}
\end{table*}

\subsection{Simulation Study in terms of Sensor Noise }
In the Earth observation satellite, the main factors that affect image quality are sensor noise and atmospheric distortions\cite{paoletti2019deep}. To conduct simulation experiments on sensor noise in satellites, we follow \cite{nalepa2021towards} and introduce three noise models: Gaussian noise, impulsive noise (salt-and-pepper noise), and Poisson noise (shot noise). The Gaussian noise simulation accounts for both thermal and quantization disturbances\cite{aggarwal2015mixed}. This type of noise is independent of the signal. Impulsive noise refers to the circumstances in which the sensor generates bad pixels during data acquisition. These bad pixels encompass both non-responsive black pixels, which fail to capture any data, and saturated white pixels\cite{tariyal2015hyperspectral}. Poisson noise is used to simulate signal-dependent photon noise due to its inherent correlation with light measurement processes\cite{rasti2018noise}.

Tables \ref{table:gaussian noise} present the classification results of the methods tested in the context of Gaussian noise, impulse noise, and Poisson noise on the Indian Pines dataset, respectively. The analysis is performed using three distinct noise scales, specifically $0.1$, $0.2$, and $0.3$. It is evident that as the noise scale increases, all tested methods exhibit considerable performance deterioration in the three noise scenarios. However, it can be observed that the proposed method continues to exhibit a relatively high level of accuracy compared to other methods in all noise scenarios, demonstrating a significant advantage. This capability is attributable to the fact that the proposed method consistently succeeds in accurately extracting analogous inter-pixel relationships by the implementation of the innovative two-stage graph construction, despite the partial disruption of spectral features. Notably, experiments conducted on impulse noise reveal that the proposed strategy, which operates independently of neighborhood information aggregation, is effective in mitigating the adverse impact of bad pixels on classification performance.

\begin{table*}[htbp]
    \centering
    \caption{AAs, OAs, and Kappa Coefficients of Different Methods on the Indian Pines Dataset with Different Level of Gaussian Noise, Impulse Noise, and Possion Noise}
    \label{table:gaussian noise}
        \resizebox{\linewidth}{!}{ 
    \begin{tabular}{@{}ccccccccccc@{}}
    \toprule
    \multirow{2}{*}{}  &\multirow{2}{*}{} &\multicolumn{3}{c}{0.1} &\multicolumn{3}{c}{0.2} &\multicolumn{3}{c}{0.3}\\ \cmidrule(l){3-11}
    & &SGL\cite{sellars2020superpixel} &STSE-DWLR\cite{zheng2019hyperspectral}  &Ours &SGL\cite{sellars2020superpixel} &STSE-DWLR\cite{zheng2019hyperspectral}  &Ours &SGL\cite{sellars2020superpixel} &STSE-DWLR\cite{zheng2019hyperspectral}  &Ours \\ \midrule
         \multirow{3}{*}{Gaussian Noise} 
         &AA &0.8086 &0.8217  &0.9164  &0.8132 &0.7520 &0.9141  &0.7672 &0.7114 &0.8987\\
         &OA &0.6993 &0.7322  &0.9656  &0.6922 &0.6556 &0.9633  &0.7672 &0.6100 &0.9450\\
         &Kappa &0.6632 &0.6976 &0.9608  &0.6571 &0.6098 &0.9582  &0.6148 &0.5570 &0.9373\\
    \midrule
    \multirow{3}{*}{Impulse Noise} 
         &AA  &0.8006 &0.7724 &0.8454  &0.7434 &0.7220 &0.7678  &0.7402 &0.6986 &0.6847\\
         &OA  &0.6838 &0.6809 &0.8933  &0.5994 &0.6270 &0.8210  &0.6005 &0.5963 &0.7571\\
         &Kappa  &0.6476 &0.6384 &0.8773  &0.5581 &0.5761 &0.7924  &0.5589 &0.5406 &0.7157\\
         \midrule
             \multirow{3}{*}{Poission Noise} 
         &AA  &0.7756 &0.7122 &0.8923  &0.7258 &0.6801 &0.8455  &0.7314 &0.6616 &0.8022\\
         &OA  &0.6768 &0.6149 &0.9389  &0.5914 &0.5772 &0.8878  &0.6246 &0.5607 &0.8214\\
         &Kappa  &0.6399 &0.5620 &0.9303  &0.5509 &0.5192 &0.8715  &0.5842 &0.4997 &0.7951\\
    \bottomrule
    \end{tabular}}
\end{table*}

\begin{table*}[htbp]
    \centering
    \caption{Ablation studies on the proposed method}
    \label{table:ablation}
    \begin{tabular}{@{}cccccc@{}}
    \toprule
    Datasets  &Metric &$w/o$ $F_{0}$ &$w/o$ top-$k$ &$w/o$ $F^{*}$ &Full \\ \midrule
         \multirow{3}{*}{Indian Pines} 
         &AA &0.9119 &0.9223 &0.8114 &0.9174 \\
         &OA &0.9550 &0.9681 &0.8338 &0.9662 \\
         &Kappa &0.9487 &0.9637 &0.8099 &0.9615 \\
    \midrule
    \multirow{3}{*}{Salinas} 
    &AA &0.9859 &0.9888 &0.9602 &0.9899 \\
    &OA &0.9783 &0.9806 &0.9331 &0.9885 \\
    &Kappa &0.9758 &0.9783 &0.9255 &0.9872 \\
    \midrule
    \multirow{3}{*}{Pavia University} 
    &AA &0.8116 &0.8571 &0.6951 &0.8256 \\
    &OA &0.8653 &0.8908 &0.7439 &0.8721 \\
    &Kappa &0.8182 &0.8537 &0.6572 &0.8251 \\
    \bottomrule
    \end{tabular}
\end{table*}

\subsection{Ablation Study}
In this subsection, an ablation study is conducted on the proposed method. The label propagation technique can be systematically categorized into three distinct phases: the initial phase involves the derivation of the preliminary soft label matrix $F_{0}$ through anchor label propagation; the subsequent phase encompasses the top-$k$-based pruning process; and the final phase involves the calculation of the final label matrix $F^{*}$ via a closed form solution. Thus, we consider three designs for the ablation study: 1) We substitute the zero matrix with $F_{0}$ as the initial label matrix. Alternatively, we revert to employing the zero matrix as the initial label matrix for the process of label propagation. 2) To enhance computational efficiency, the sparse graph is employed. However, this approach may incur a reduction in accuracy. Therefore, the fully connected graph is used instead of sparse graphs to investigate the impact of this potential loss on the overall classification accuracy. 3) The utilization of sparse graph-based label propagation enhances the exploitation of inherent correlations among unclassified pixels, thus refining the initial classification labels. As an alternative, we omit the sparse graph-based label propagation. In this case, we investigate the extent to which overall classification accuracy can be maintained by employing rapid label propagation via the anchor graph. Table \ref{table:ablation} presents the results of the ablation studies conducted on the three evaluated datasets. It is evident that a certain level of performance degradation is observed across all three datasets upon the removal of $F_{0}$ or $F^{*}$. This substantiates the effectiveness of the proposed two-stage label propagation method. The use of a fully connected graph instead of the pruned sparse graph may result in a marginal enhancement in classification performance. Nevertheless, this improvement is not substantial when compared to the classification performance achieved through sparse graphs. In contrast, the adoption of a fully connected graph leads to an exponential escalation in computational time. Consequently, with respect to the algorithm's overall computational efficiency, the sparse graph is deemed to be more suitable.

\subsection{Parameter Sensitivity}
\begin{figure*}[htbp]
	\centering
	\subfigure[Indian Pines]{
		\begin{minipage}[t]{0.31\linewidth}
			\centering
			\includegraphics[width=0.99\hsize, height=0.6\hsize]{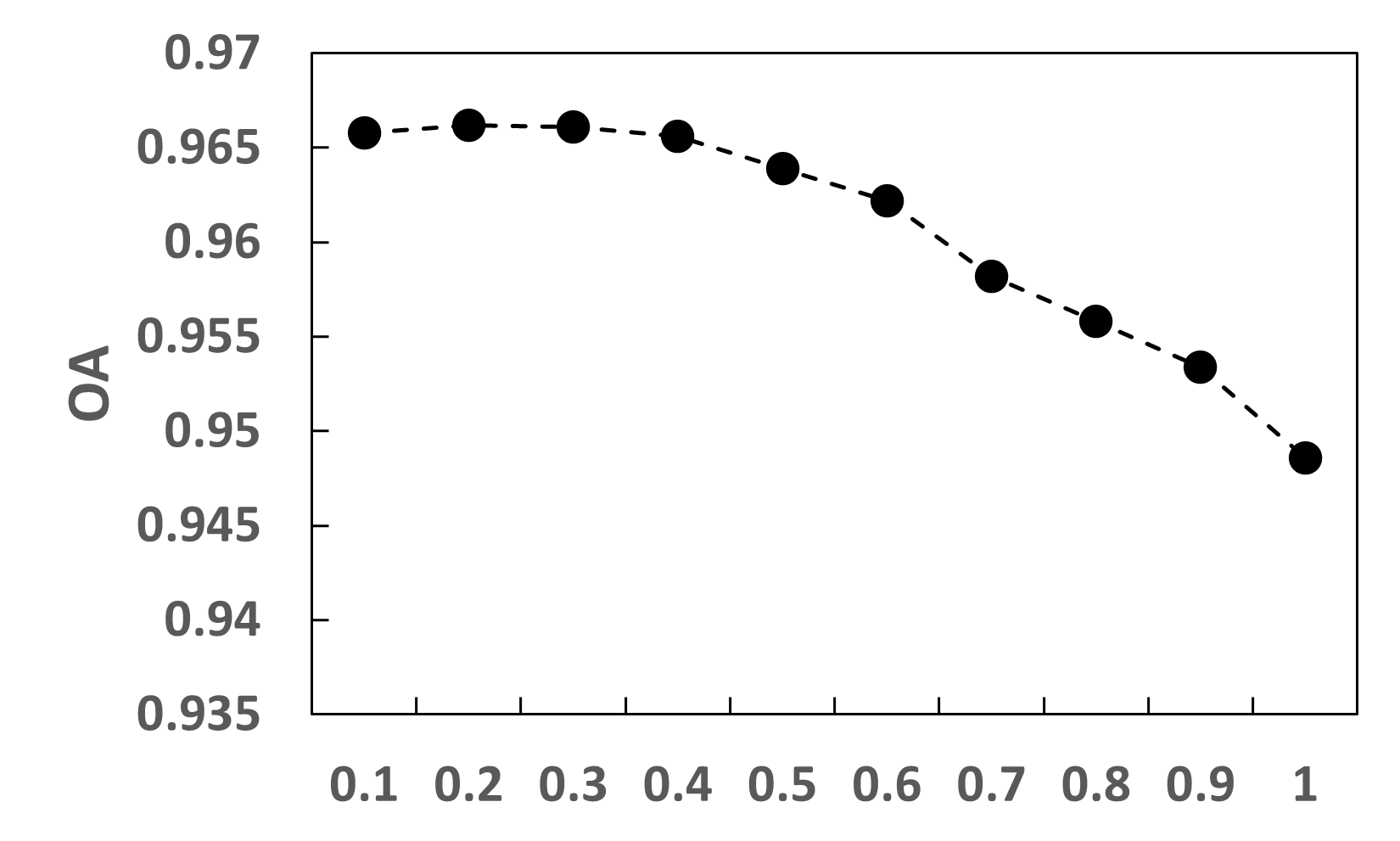}\\
   \end{minipage}}
   	\subfigure[Salinas]{
		\begin{minipage}[t]{0.31\linewidth}
			\centering
			\includegraphics[width=0.99\hsize, height=0.6\hsize]{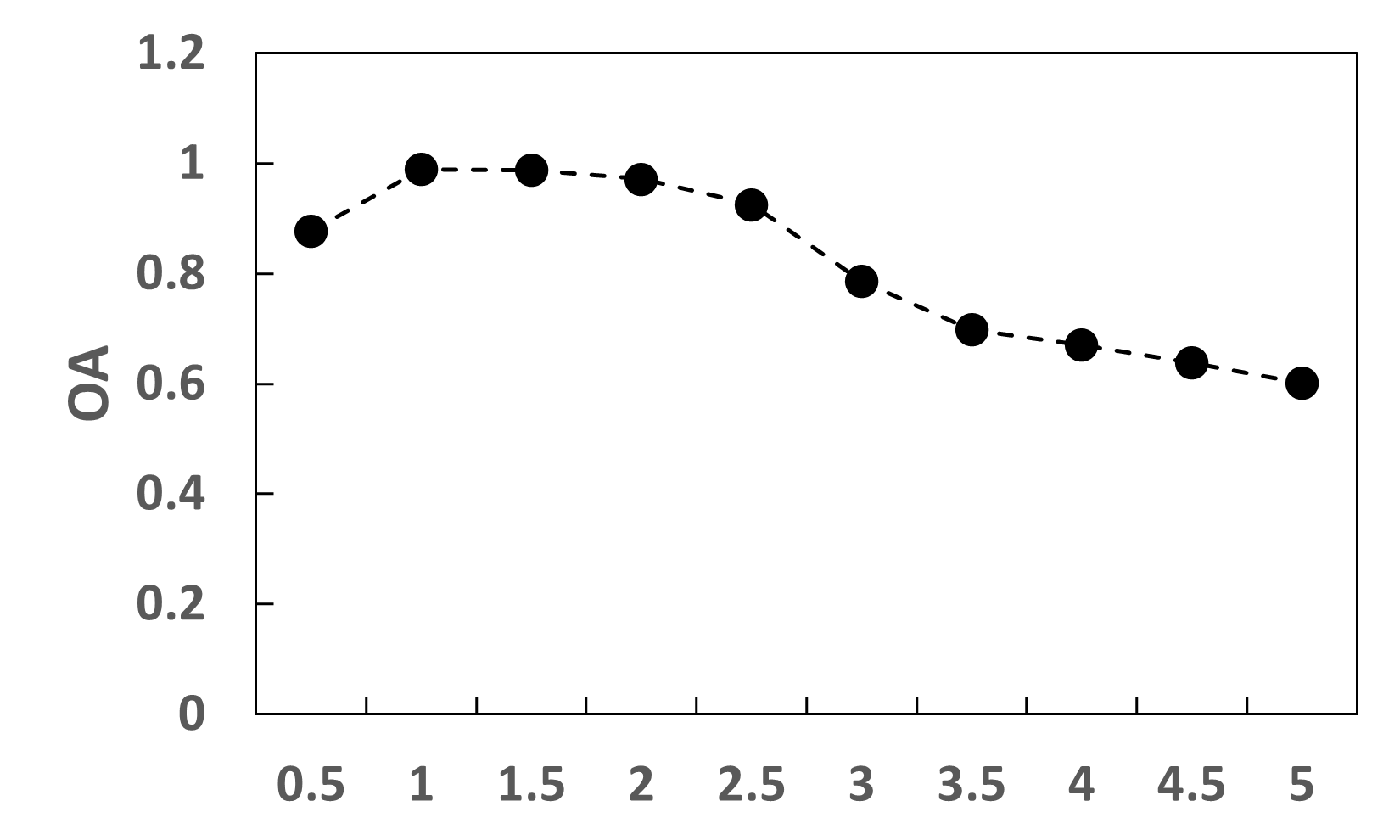}\\
			\end{minipage}}
   	\subfigure[Pavia University]{
		\begin{minipage}[t]{0.31\linewidth}
			\centering
			\includegraphics[width=0.99\hsize, height=0.6\hsize]{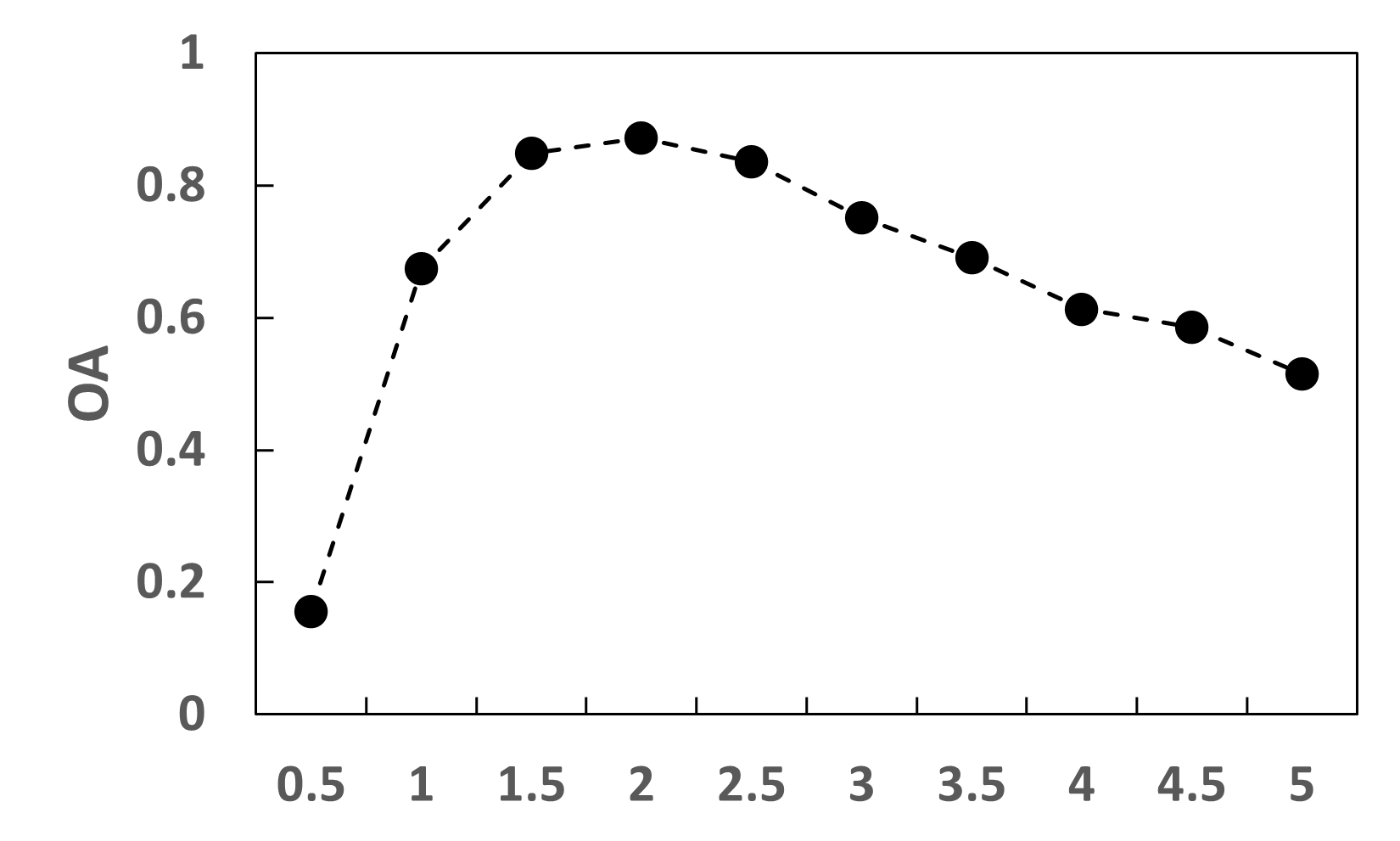}\\
			\end{minipage}}
	\centering
	\caption{Sensitivity analysis on the value of $\sigma^{2}$ on the Indian Pines, Salinas, and Pavia University datasets. (a) Indian Pines. (b) Salinas. (c) Pavia University.}
	\label{fig:sigma}
\end{figure*}

\begin{figure}[htbp]
	\centering
	\includegraphics[width=0.9\linewidth]{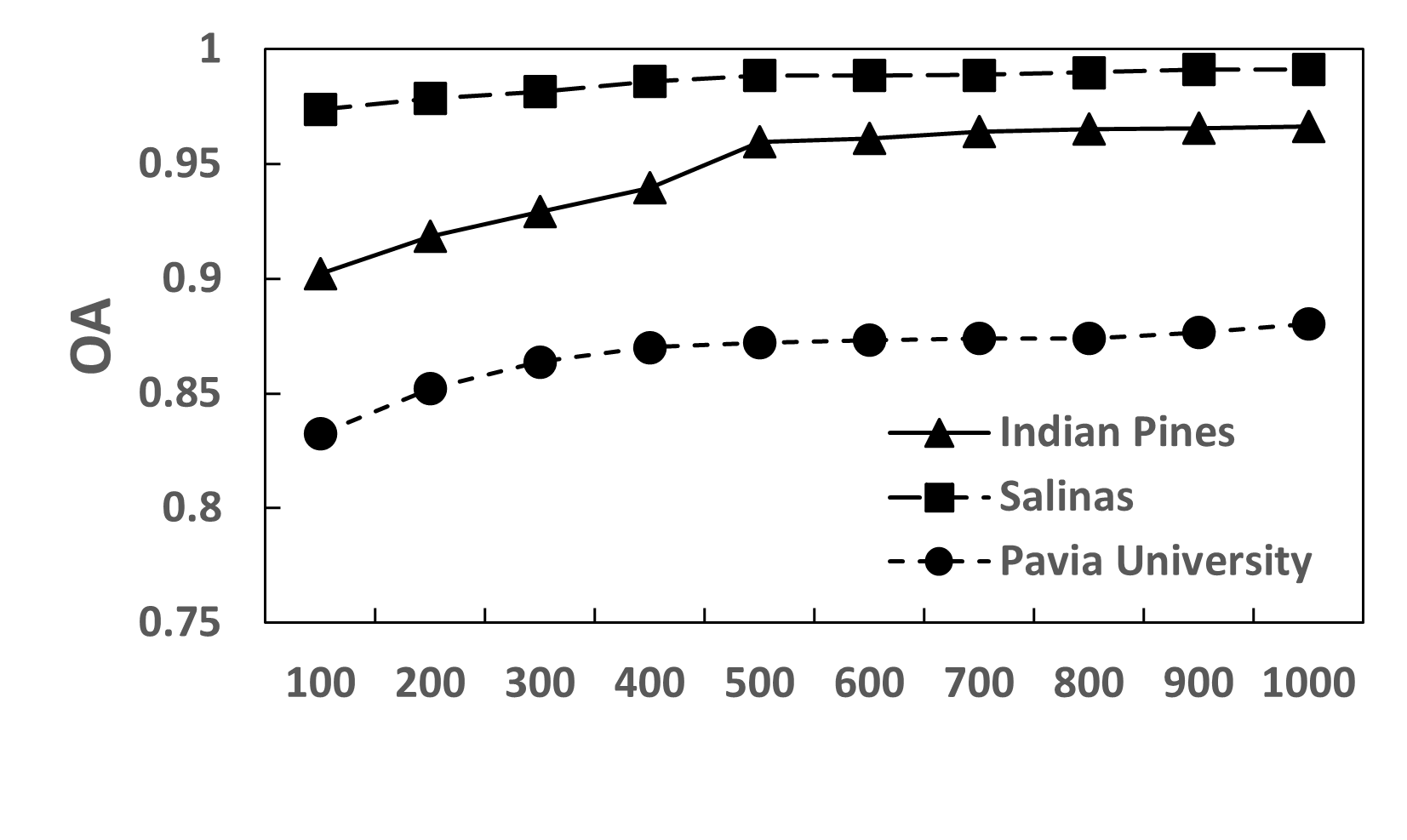}
	\caption{Sensitivity analysis on the value of $k$ on the Indian Pines, Salinas, and Pavia University datasets.}
	\label{fig:k}
\end{figure}

In this subsection of the experiments, we investigate the sensitivity of the proposed method with respect to the parameter configurations. The parameter settings are divided into two distinct categories: the sensitivity analysis of parameters for the label propagation phase and the sensitivity analysis for the parameters related to the clustering phase. Two critical parameters must be manually configured during the label propagation process: the bandwidth $\sigma$ of the Gaussian kernel function used in graph construction and the value of $k$ required for selection in the pruning stage. The bandwidth $\sigma$ is a fundamental parameter of the Gaussian kernel function, which plays a crucial role in defining the smoothness of the function. Fig.\ref{fig:sigma} illustrates the impact of different $\sigma$ on the overall accuracy of the classification. Notably, within the Salinas and Pavia University datasets, distinct inflection points can be identified. Therefore, for the Salinas and Pavia University datasets, the values of $\sigma^{2}$ are specified as $1$ and $2$, respectively. In the case of the Indian Pines dataset, the maximum classification precision is reached at a $\sigma^{2}$ value of $0.2$, with a subsequent decrease as the $\sigma$ value increases. Consequently, the optimal value of $\sigma^{2}$ for the Indian Pines dataset is determined to be $0.2$. The selection of the parameter $k$ during the pruning process significantly influences not only the accuracy of the classification but also the computational efficiency of the algorithm. Fig.\ref{fig:k} illustrates the impact of varying the selections of the parameter $k$ on the performance of the proposed method in the three datasets. In general, the classification accuracies in all three datasets exhibit an increasing trend as the parameter $k$ increases. However, this enhancement becomes marginal as $k$ exceeds a particular threshold. Therefore, to maintain an equilibrium between computational efficiency and classification accuracy, the parameter $k$ is configured as $1000$, $500$, and $500$ for the Indian Pines, Salinas, and Pavia University datasets, respectively. 

\begin{figure}[htbp]
        \centering
	\includegraphics[width=0.9\linewidth]{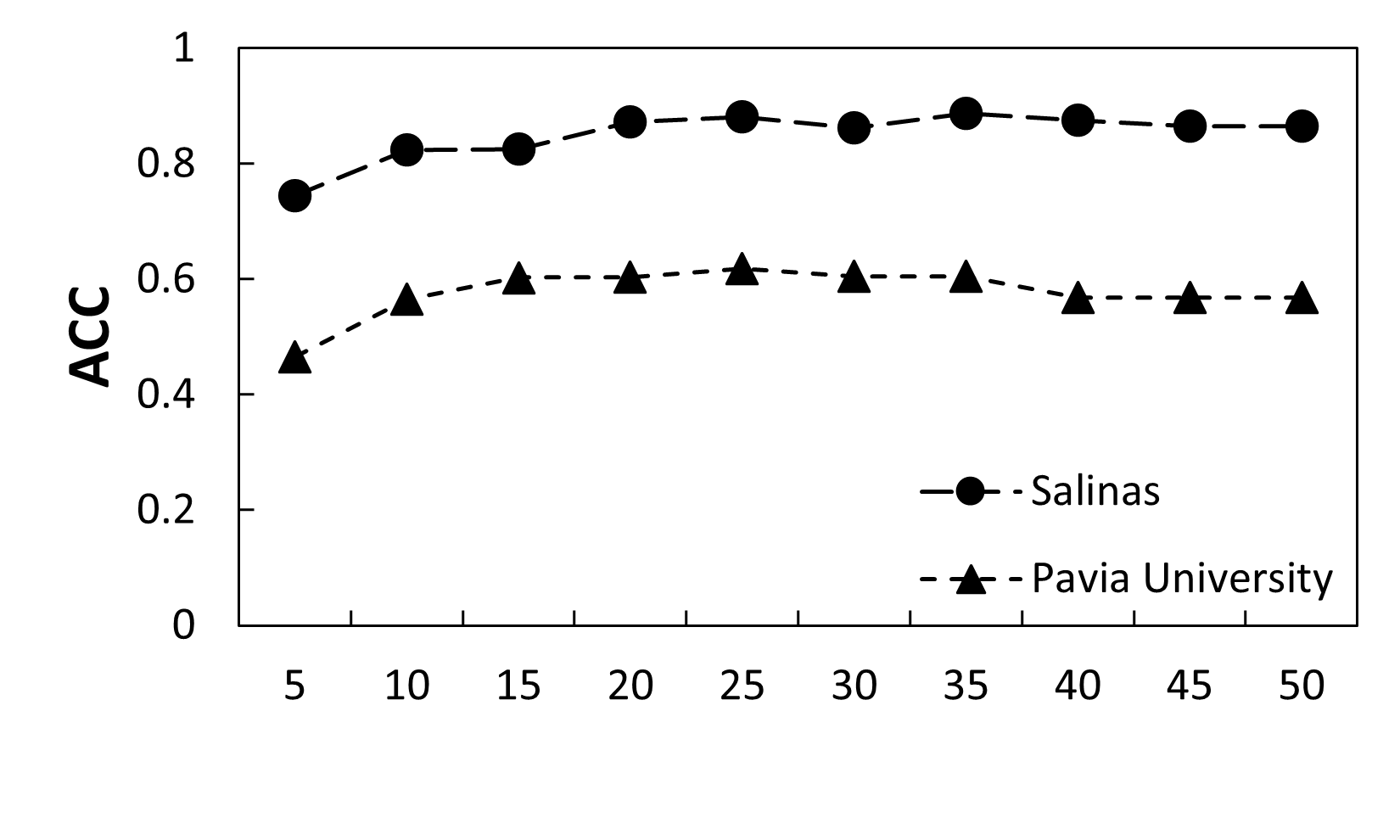}
	\caption{Sensitivity analysis on the value of $\beta$ on the Salinas and Pavia University datasets.}
	\label{fig:beta}
\end{figure}

\begin{figure*}[htbp]
	\centering
   	\subfigure[]{
		\begin{minipage}[t]{0.48\linewidth}
			\centering
			\includegraphics[width=0.99\hsize, height=0.6\hsize]{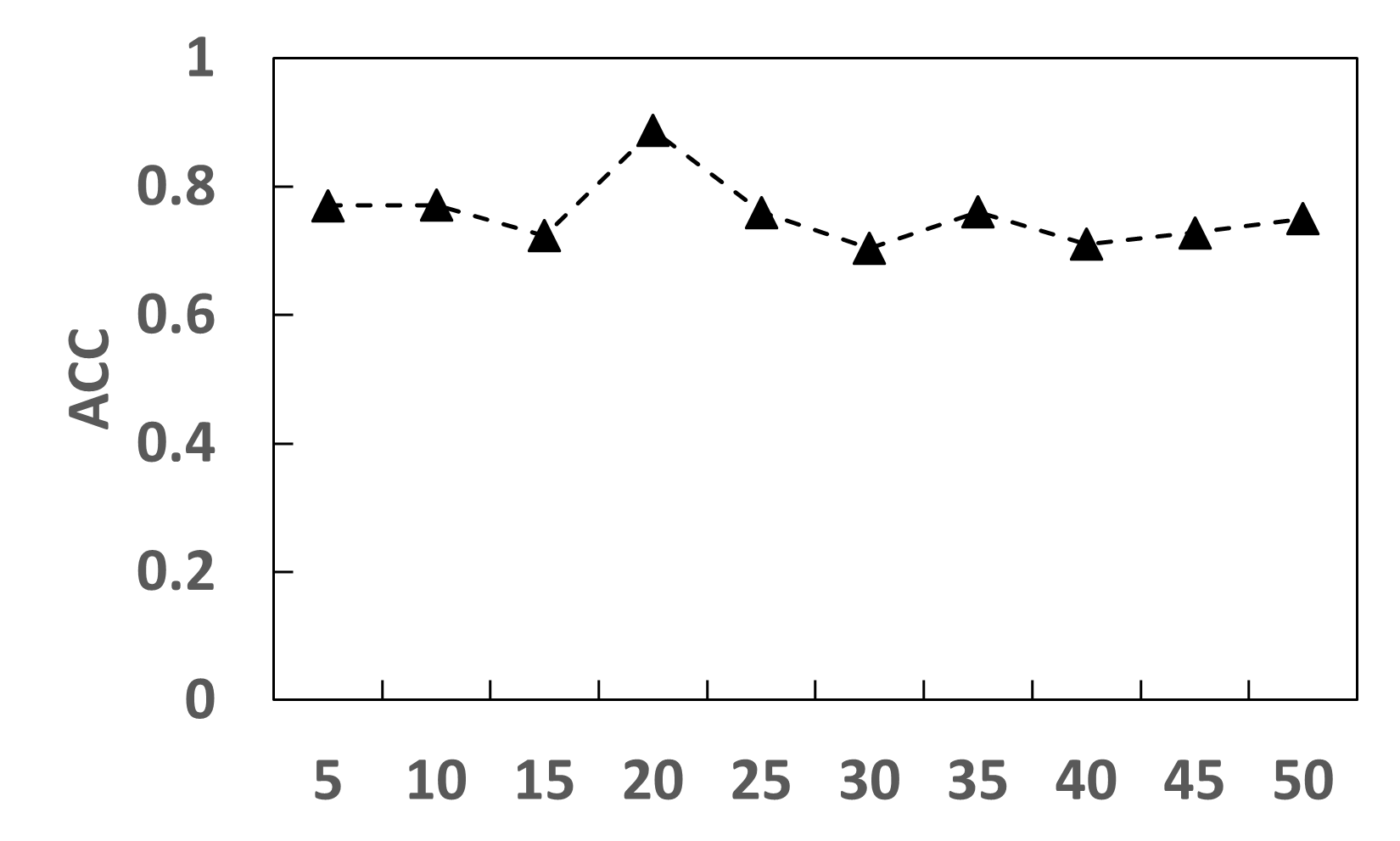}\\
			\end{minipage}}
	\centering
       	\subfigure[]{
		\begin{minipage}[t]{0.48\linewidth}
			\centering
			\includegraphics[width=0.99\hsize, height=0.6\hsize]{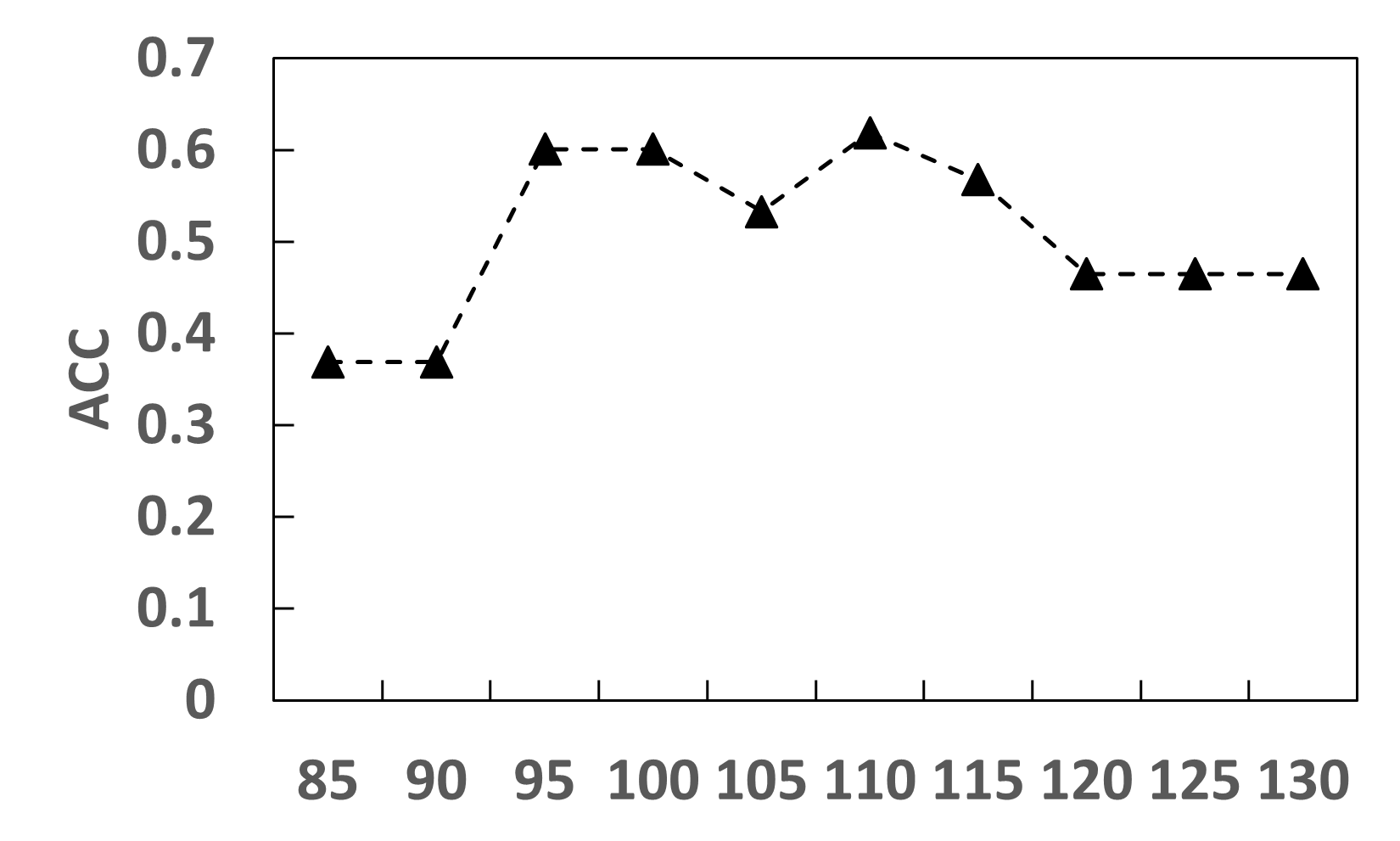}\\
			\end{minipage}}
	\centering
	\caption{Sensitivity analysis on the value of $h$ on the Salinas and Pavia University datasets. (a) Salinas. (b) Pavia University.}
	\label{fig:h}
\end{figure*}

In the proposed clustering algorithm, there exist two pivotal parameters that require manual adjustment: the first is $\beta$, and the second is the number of nonzero terms $h$. Fig.\ref{fig:beta} illustrates the changes in the precision of the clustering under different $\beta$ values. In general, as $\beta$ increases, the effect of rank constraints becomes more pronounced, and the clustering effect eventually stabilizes. The values for the first parameter are set to $35$ for the Salinas dataset and $25$ for the Pavia University dataset, respectively. Regarding the value of $h$, an excessively large $h$ is likely to result in the accuracy being compromised by noise; conversely, an excessively small $h$ may lead to inadequate feature information for effective clustering. Consequently, in the classification results for $h$ as illustrated in Fig.\ref{fig:h}, a distinct peak is observable. Finally, we set the $h$ values for the Salinas and Pavia University dataset at $25$ and $110$, respectively.

\subsection{Onboard Computational Efficiency Analysis}
To evaluate the time efficiency of the proposed method and its applicability for onboard deployment, we conducted a time analysis between the proposed method and other machine learning algorithms including K-Nearest Neighbors (KNN), $k$-means, Support Vector Machine (SVM), Random Forest (RF), and Fuzzy C-Means (FCM). Table \ref{table:time} shows the inference times for each evaluated method, where Ours-G and Ours-O represent the inference time of the proposed method on the ground processing platform and the simulation onboard processing platform, respectively. The ground processing platform utilizes an Intel Core i7 9700 CPU. It is observed that, when compared to machine learning algorithms such as KNN, k-means, SVM, and RF, the proposed method does not exhibit an advantage. However, considering the huge gap in classification accuracy, the time disadvantage of the proposed method is not unacceptable. In particular, by reducing the parameter $k$ to a value of $10$, the proposed method achieves a reduction in the ground inference time that exceeds two-thirds, while maintaining an overall classification accuracy comparable to that of the state-of-the-art method. To test the onboard inference speed of the proposed method, we simulated the hardware environment on the satellite. HYPSO-1 \cite{langer2023robust} is chosen as the reference model. This small hyperspectral imaging satellite employs a dual-core Cortex-A9 CPU to construct its data processing platform. To validate the computing speed of the proposed method on a satellite, the clock speed and core count of the ground platform CPU are restricted to $1$ GHz and $2$ cores, respectively, aligning with the specifications of the Cortex-A9. Although the simulated onboard environment demonstrates an increase in inference time compared to ground-based computations, it still meets the requirements for fast processing in onboard hyperspectral image classification tasks.
\begin{table*}[htbp]
    \centering
    \caption{Runing time comparison (in seconds) among different methods}
    \label{table:time}
    \begin{tabular}{@{}cccccccc@{}}
    \toprule
    &KNN &$k$-means &SVM &RF &FCM &Ours-G &Ours-O\\ \midrule
    Indian Pines &0.54 &0.77 &0.19 &28.99 &55.38 &31.03 &108.64\\
    Salinas &2.08 &2.73 &1.01 &85.34 &480.28 &175.51 &657.30\\
    Pavia University &1.62 &2.23 &0.63 &50.04 &2348.66 &133.52 &485.31\\
    \bottomrule
    \end{tabular}
\end{table*}

\section{Discussion}

\subsubsection*{Special Requirements for In-Orbit Hyperspectral Image Classification}
In-orbit hyperspectral image classification fundamentally differs from conventional ground-based offline processing. Due to the stringent constraints imposed by the low-power and limited computational capacity of satellite onboard processors, the in-orbit environment is generally unable to support complex deep neural network inference. Consequently, classification models must adopt lightweight architectures with low computational complexity and high deployability on resource-limited hardware platforms. Moreover, restricted onboard storage and downlink bandwidth require that in-orbit classification be executed in real time during data acquisition to prevent data accumulation and ensure the timeliness of subsequent observation tasks or autonomous control operations. In addition, satellite-borne hyperspectral sensors are susceptible to band noise, dead pixels, and spectral drift arising from long-term exposure to space radiation, thermal fluctuations, and attitude disturbances. These factors significantly degrade data quality relative to ground-calibrated conditions, thereby imposing stringent requirements on the model’s noise robustness and its ability to generalize across temporal and environmental variations. Furthermore, collecting high-quality and spatially extensive annotated data is inherently challenging, and frequent domain shifts occur across different orbits, seasons, and surface types. Therefore, in-orbit hyperspectral image classification must concurrently address the demands for real-time performance, high robustness, strong generalization, and operational reliability, under stringent computational constraints. In this context, our method exhibits strong advantages in noise robustness, few-shot learning, and applicability in non–deep learning settings. However, due to the inherent time complexity of graph-based processing, its real-time performance remains slightly limited. In subsequent work, we plan to further reduce the number of edges in graph by optimizing pruning strategies to achieve graph sparsification and enhance computational speed.

\subsubsection*{Selection and Generation of Initial Anchor Labels}
The quality and distribution of the initial labels exert a critical influence on the final classification outcomes in label propagation algorithms. Biased or erroneous initial labels tend to be amplified through the graph structure, subsequently degrading the overall classification performance. Moreover, adequate coverage of the initial labels within the feature space is essential. When the initial annotations are overly sparse or unevenly distributed across the data manifold, the propagation process may fail to reach all samples, leading to unstable local decisions and indistinct class boundaries. In our experiments, under conditions of extreme anchor scarcity, we found that classes with very limited sample proportions were occasionally omitted during the sampling process. This issue arises because the anchor selection strategy we used trades off a degree of category discrimination to meet the demand for fast response. For instance, in the Indian Pines dataset, Class $9$ contains only $20$ samples, whereas Class $11$ comprises $2,455$ samples, reflecting a pronounced class imbalance. when the number of anchor was set to a extremely low value of $5$ per class, we observed that no samples from Class $9$ were selected during the anchor sampling procedure, ultimately resulting in classification failures for this minority class. The issue was mitigated when the number of anchors was increased to $10$ per class, ensuring coverage of all pixel classes and leading to an improvement in overall classification accuracy of 1.82\%.

During in-orbit operations, satellites continuously pass over unseen regions, and it is infeasible to upload annotated samples in real time via satellite–ground links. Consequently, autonomously generating labels for initial anchors becomes a central challenge. A common solution is to pre-load a spectral library containing reference reflectance curves acquired under controlled laboratory conditions. Although spectral matching can provide approximate semantic labels, such libraries often fail to capture seasonal variations in vegetation or environmental fluctuations in water bodies, and are highly susceptible to errors caused by noisy or mixed pixels. More critically, spectral libraries are inherently static and cannot adaptively infer the number of classes or underlying semantic structures, substantially limiting their applicability in diverse scenarios. To overcome these limitations, we choose to generate pseudo labels for anchors using unsupervised clustering without relying on prior knowledge. This approach does not require a pre-built spectral library or ground-based annotation. Based on the assumption of spectral homogeneity, pixels of the same land-cover type are expected to form compact clusters in feature space. Compared with spectral library matching, this strategy offers greater adaptability and robustness for in-orbit hyperspectral image classification across varying conditions.

\subsubsection*{Probabilistic Label Learning Against Mixed Pixels}
Given the limited spatial resolution and high spectral dimensionality of spaceborne hyperspectral imagery, individual pixels frequently contain multiple land-cover components, resulting in mixed pixels. The presence of mixed pixels substantially exacerbates intra-class spectral variability while diminishing inter-class separability, thereby increasing the intrinsic difficulty of hyperspectral image classification. Contemporary deep learning–based classifiers typically depend on large quantities of clean and label-consistent training samples. However, mixed pixels introduce ambiguity and label noise that can significantly impair the performance of such models, particularly in boundary regions where class mixing is most prevalent. In contrast to conventional deep learning approaches that rely on hard labels, our method employs a soft-label propagation mechanism, enabling more robust and reliable classification in the presence of spectral mixing. In the first stage, initial soft labels for all samples are generated via anchor-based label propagation. In the second stage, these preliminary soft labels are iteratively refined to obtain the final label–probability distributions. Throughout the entire framework, soft labels are consistently employed to represent label information, thereby facilitating the effective utilization and propagation of the probabilistic attributes associated with each sample.

\section{Method}
\subsection{Overview}
\begin{figure*}[htbp]
	\centering
	\includegraphics[width=1\linewidth]{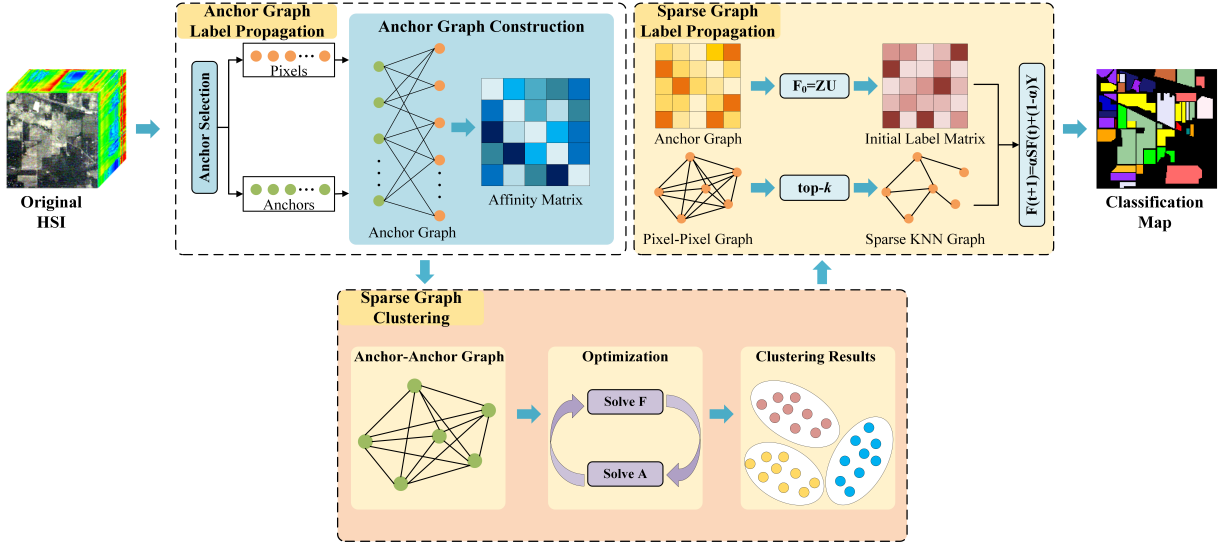}
	\caption{Overall architecture of the proposed method.}
	\label{overview}
\end{figure*}

Given a hyperspectral dataset $X=\{x_{i}|i=1,...,N,x_{i}\in \mathbb{R}^{d}\}$ and a class set $C=\{1,...,c\}$, where $d$ and $c$ represent the dimension of the spectral band and the class of labels, respectively, the label set corresponding to the hyperspectral image dataset $X$ is $L\in \mathbb{R}^{N}$, where the first $l$ pixels are labeled and the remaining $u$ are unlabeled. The goal of hyperspectral image classification is to obtain the label matrix $F=[F_{0}^{T},...,F_{N}^{T}]^{T}\in \mathbb{R}^{N\times c}$. In this section, we will derive the specific steps of the proposed onboard hyperspectral image classification method. First, we will show how to construct an anchor graph and introduce the principle of label propagation based on the anchor graph in Section \ref{subsection:anchor graph}. In Section \ref{subsection:KNN Graph}, an approach to obtaining the closed-form solution on the sparse graph is used as an alternative to the iterative process of label propagation, improving computational efficiency. Moreover, in Section \ref{subsection:cluster}, the proposed clustering algorithm is discussed in further detail to derive the anchor label to meet the requirements onboard. The comprehensive algorithmic process is illustrated in Algorithm \ref{alg:1} and Fig.\ref{overview}.

\begin{algorithm}
	\renewcommand{\algorithmicrequire}{\textbf{Input:}}
	\renewcommand{\algorithmicensure}{\textbf{Output:}}
        \caption{The algorithm of proposed method.}
	\label{alg:1}
	\begin{algorithmic}[1]
            \REQUIRE Hyperspectral dataset $X\in \mathbb{R}^{(m+n)\times d}$, the number of anchors $m$, the width of Gaussian kernel function $\sigma$, the number of nearest neighbors $k$, balance parameter $\alpha$.
            \ENSURE The label matrix $F\in \mathbb{R}^{(m+n) \times c}$.
            \STATE Generate $m$ anchors $Q \in \mathbb{R}^{m\times d}$ and obtain the corresponding label set of anchors $U \in \mathbb{R}^{m \times c}$;
            \STATE Construct the anchor graph $Z\in \mathbb{R}^{n\times m}$ by Eq.\ref{eq:1};
            \STATE Calculate the initial label matrix $F_{0}\in\mathbb{R}^{n \times c}$ and affinity matrix $W^{a}\in\mathbb{R}^{n \times n}$ based on anchor graph $Z$ for unlabeled pixels by Eq.\ref{eq:2} and Eq.\ref{eq:3};
            \STATE Perform top$-k$ pruning strategy to construct the sparse KNN graph $W^{k}\in\mathbb{R}^{n \times n}$ for unlabeled pixels;
            \STATE Construct the affinity matrix $W\in\mathbb{R}^{(n+m) \times (n+m)}$ for all data samples by Eq.\ref{eq:6};
            \STATE Derive the similarity matrix $S\in\mathbb{R}^{(n+m) \times (n+m)}$ using the closed-form solution provided in Eq. \ref{eq:8};
            \STATE Calculate the final label matrix $F\in\mathbb{R}^{(n+m) \times c}$ by Eq.\ref{eq:12}.
	\end{algorithmic}  
\end{algorithm}

\subsection{Anchor Graph Based Label Propagation}
\label{subsection:anchor graph}
Conventional graph-based semi-supervised approaches\cite{zhang2019label}\cite{sellars2020superpixel}\cite{jia2020superpixel}\cite{zhong2021dynamic} for hyperspectral image classification generally employ a fixed number of pixels extracted from each predefined land cover category as labeled samples. However, these approaches require prior knowledge of the land cover category in the observation area, which can be a burden in numerous scenarios. To address this issue, we consider the development of a method to uniformly sample all categories without requiring prior knowledge of the land cover category. Anchors are the most representative points in a given dataset, which can well indicate the overall distribution of the data and the internal structure information. Through the generation of anchors for a given hyperspectral image dataset, it becomes feasible to uniformly extract samples from all land cover categories without prior knowledge. The selected anchors can be subsequently labeled by manual annotation or unsupervised techniques to obtain label information. The method for acquiring anchor labels through unsupervised techniques will be elaborated in Section \ref{subsection:cluster}. The construction of the anchor graph can be divided into two phases: the first phase entails the generation of anchors, and the second involves the computation of the similarity between anchors and samples. 

For anchor generation, two primary approaches are used: the random selection strategy and the $k$-means strategy. The selection of anchors is vital for the propagation of the label. Although the random selection strategy is notably efficient, it does not guarantee comprehensive coverage of all types of land cover by the generated anchors. In contrast, the $k$-means strategy procures more representative points, thereby improving the performance of the overall algorithm. Considering that the clustering centers derived from the $k$-means algorithm may not be present within the provided dataset, the nearest points are calculated and designated as anchors.

Given the set of generated anchors $Q=\{q_{i}|i=1,...,m, q_{i} \in \mathbb{R}^{d}\}$ and the set of unlabeled pixels $P=\{p_{i}|i=1,...,n, p_{i} \in \mathbb{R}^{d}\}$, we need to construct the anchor graph $Z\in \mathbb{R}^{n \times m}$ between pixels and anchors. We used the Gaussian kernel function to serve as a metric to assess the similarity between the $i_{th}$ pixel and the $j_{th}$ anchor. The corresponding formula is demonstrated as follows:
\begin{align}
\label{eq:1}
    z_{ij}=exp (\frac{-\left \|x_{i}-a_{j}\right \|_{2}^{2}}{2\sigma^{2}} )
\end{align}
where $\delta$ is the band width. Due to $m<<n$, the calculation of the anchor graph $Z$ is executed with high efficiency. Subsequent to the obtain of the anchor graph $Z$ and the anchor label $U \in \mathbb{R}^{m \times c}$, the initial label matrix $F_{0}\in \mathbb{R}^{n \times c}$ for all pixels can be efficiently determined by label propagation\cite{nie2023fast}:
\begin{align}
\label{eq:2}
    F_{0}=ZU
\end{align}
The information of the anchor label is distributed to all pixels by exploiting the similarity relationship between the anchors and the pixels. Moreover, the initial affinity matrix $W_{1}\in \mathbb{R}^{n \times n}$ between pixels can be deduced by utilizing the anchor graph $Z$, which is as follows\cite{liu2010large}:
\begin{align}
\label{eq:3}
    W^{a}=Z\Delta^{-1} Z^{T}
\end{align}
where $\Delta \in \mathbb{R}^{m \times m}$ is a diagonal matrix and the $i_{th}$ entry is
defined as $\sum^{n}_{j=1}z_{ij}$. It is evident that $W_{a}$ constitutes a stochastic matrix.

\subsection{Sparse Graph Based Label Propagation}
\label{subsection:KNN Graph}
Despite having derived the initial pixel label $F_{0}$, they are still not precise enough due to the absence of a direct assessment of the similarity relationship between pixels. However, estimating pairwise similarity among all pixels exhibits a time complexity of $O(n^{2})$, which demands substantial computational resources. Such an approach is undesirable. Therefore, we opt to construct a sparse graph. 

The prevailing approach in constructing the sparse graph entails a preliminary calculation of similarity across all pairs of samples, followed by the extraction of the k-nearest neighbors for each individual sample. However, this practice is evidently very time consuming, especially for the pixel-wise framework. Thus, a pre-pruning top-$k$ strategy is devised to circumvent the need for estimating similarity among all possible pairs of pixels. By applying the pruning strategy, a sparse graph $W^{k} \in  \mathbb{R}^{n \times n}$ can be derived. Specifically, for each pixel, we retain the $k$ edges corresponding to the largest similarity and eliminate other connections. In this section, the initial affinity matrix $W^{a}$ is used as the fully connected graph to be pruned. The edge set $\varepsilon$ of the pruned sparse graph $W^{k}$ is as follows:
\begin{align}
    \varepsilon_{i}=\{(i,j)|j=\underset{j\neq i}{topk(W^{a}_{i})}\} 
\end{align}
where $\varepsilon_{i}$ represents the edge set of $i_{th}$ pixel and $W^{a}_{i}$ means the $i_{th}$ row of $W^{a}$. After the pruning phase, the similarity between the pixel pairs is recomputed as follows.
\begin{align}
    w^{k}_{ij} =
    \begin{cases}
        exp (\frac{\left \|x_{i}-x_{j}\right \|_{2}^{2}}{2\sigma^{2}} &if \space x_{j} \in N(x_{i})\\
        0 &otherwise
    \end{cases}
\end{align}
where $N(x_{i})$ represents the pixel set of the $k$ nearest neighbors of $x_{i}$.

In Eq.\ref{eq:2}, the label information is propagated exclusively from the anchors to the pixels. In addition to the similarity relations between anchors and pixels, it is imperative to fully exploit the internal similarity relations between pixels in label propagation. Firstly, we establish the affinity matrix $W\in \mathbb{R}^{(m+n) \times (m+n)}$ for all samples, including labeled anchors and unlabeled pixels. Hence, the affinity matrix $W$ can be divided into four submatrixes:
\begin{align}
\label{eq:6}
	W=
	\left[
	\begin{array}{cc}
	W_{ll}& W_{lu} \\  
	W_{ul}& W_{uu}
	\end{array}
	\right]
\end{align}
where $W_{ll} \in \mathbb{R}^{m\times m}, W_{uu} \in \mathbb{R}^{n \times n}, W_{lu}=Z, W_{ul}=Z^{T}$. The anchor-anchor graph $W_{ll}$ encapsulates the similarity relations among the anchors. Given the limited number of anchors, we choose to construct a fully connected adjacency matrix. For the pixel-pixel graph $W_{uu}$, we integrate the affinity matrix $W^{a}$ with $W^{k}$ and resolve it in a subsequent way:
\begin{align}
    W_{uu}=W^{a} \cdot W^{k}
\end{align}
where $\cdot$ represents element-wise product. To serve as a foundation for iterative propagation, the affinity matrix $W$ is symmetrically normalized:
\begin{align}
\label{eq:8}
    S&=D^{-1/2}WD^{-1/2}
\end{align}
where $D$ is the degree matrix of $W$:
\begin{align}
    D =
    \begin{cases}
        \sum^{(m+n)}_{j=1}w_{ij} &i= j\\
        0 &i\neq j
    \end{cases}
\end{align}

Typically, the label set $L$ can define a matrix $Y\in \mathbb{R}^{(m+n)\times c}$ with $Y_{ij}=1$ provided that the sample $x_{i}$ is classified as class $j$; otherwise, $Y_{ij}=0$. Let $Y$ represent the label matrix to be propagated and denote $F\in R^{(m+n)\times c}$ as the derived label matrix after the label propagation. The prediction for unlabeled pixels can be represented through the following dynamic process:
\begin{align}
\label{eq:10}
    F(t+1)=\alpha SF(t)+(1-\alpha)Y
\end{align}
where $t$ represents the $t_{th}$ iteration and $F(0)=Y$. $\alpha$ is the balance parameter utilized to acquire both the label information from $Y$ and the historical iterations. In every iteration, each pixel acquires information from its connected samples while simultaneously preserving a part of its initial information from $Y$. However, the label matrix $Y$ exhibits a limitation in which the initial label of the unlabeled pixels is assumed to be $0$, indicating that the matrix comprising the final $m$ rows of Y is a zero matrix. Consequently, in each iteration, these pixels are ascribed a value of $0$ as their own information, which is evidently untenable. To address this limitation, we integrate the anchor label $U \in 
 R^{m\times c}$ with the initial pixel label $F_{0} \in 
 R^{n\times c}$, forming the initial label matrix. The proposed label matrix $Y$ is as follows:
 \begin{align}
     Y=concat(U, F_{0})
 \end{align}
where $concat(\cdot)$ represents the operation of concatenation. Finally, a label $l_{i}$ is attributed to each sample $x_{i}$ according to the rule of $l_{i}=\mathop{\mathrm{argmax}}_{j \leq c}{F_{ij}}$. In addition, the research in \cite{zhou2003learning} asserts the existence of a closed form solution for Eq.\ref{eq:10}. They have shown that with an increasing number of iterations, the optimal $F^{*}$ approaches a constant value only dependent on $Y$:
\begin{align}
\label{eq:12}
    F^{*}=(I-\alpha S)^{-1}Y
\end{align}
The closed-form solution eliminates the need for a lengthy iterative process. The more detailed proof can be found in \cite{zhou2003learning}.

\subsection{Sparse Graph-based Clustering}
\label{subsection:cluster}
The preceding section elaborates on the application of a semi-supervised framework to facilitate the fast classification of hyperspectral images collected from an observation area. However, the acquisition of anchor label remains an unresolved challenge. The conventional procedure involves the manual annotation of anchors after the transmission of hyperspectral images to the ground. However, it fails to satisfy the fast response requirement. To address this issue, a graph-based clustering method is proposed in this section to determine the anchor label.

\begin{theorem}
\label{theorem:1}
\textit{The multiplicity $c$ of the eigenvalue zero of the Laplacian matrix $L_{A}$ is equal to the number of connected components in the graph associated with $A$.} 
\end{theorem}

Having acquired the affinity matrix $W_{ll}$ among the anchors, we hereby impose the rank constraint\cite{nie2016constrained} on the anchor-anchor graph $W_{ll}$ to learn a sparse similarity matrix $A$ with $c$ connected components, where $c$ is the number of classes. Compared to the original anchor-anchor graph $W_{ll}$, $A$ is a more suitable target for clustering. Meanwhile, using the reconstructed sparse graph $A$ to replace the original fully connected anchor-anchor graph $W_{ll}$ in the second stage of label propagation can further improve computational efficiency and classification accuracy. Theorem \ref{theorem:1} elucidates the correlation between the rank constraint and the number of connected components in a graph\cite{mohar1991laplacian}. Consequently, the challenge of constructing a graph with $c$ connected components can be reformulated as learning a similarity matrix $A$ that exhibits $rank(L_{A})=m-c$. The objective function with rank constraint to be optimized is as follows:
\begin{align}
\label{eq:13}
    &\mathop{\min}_{A} \left \| A-W_{ll} \right \|^{2}_{F} + \gamma\left \| A \right \|^{2}_{F}\\ \notag
& \begin{array}{r@{\quad}l@{}l@{\quad}l}
s.t.& \sum_{j}{A_{ij}=1}, A_{ij}>0, rank(L_{A})=m-c.\\ 
\end{array}
\end{align}
where the inclusion of the second term facilitates the sparsification of the graph $A$. Due to the inherent complexities involved in directly addressing the rank constraint, it is accordingly relaxed into an eigenvalue constraint. Given that $L_{A}$ is a positive semi-definite matrix, it follows that the eigenvalues of $L_{A}$ are all nonnegative. Order the eigenvalues in ascending sequence and let $\lambda_{i}(L_{A})$ denote the $i_{th}$ smallest eigenvalue. Eq.\ref{eq:13} can be transformed into:
\begin{align}
    &\mathop{\min}_{A} \left \| A-W_{ll} \right \|^{2}_{F}+ \gamma\left \| A \right \|^{2}_{F}+2\beta \sum_{i=1}^{c}\lambda_{i}(L_{A})\\ \notag
& \begin{array}{r@{\quad}l@{}l@{\quad}l}
s.t.& \sum_{j}{A_{ij}=1}, A_{ij}>0.\\ 
\end{array}
\end{align}
Based on Ky Fan's Theorem\cite{fan1949theorem}, the eigenvalue constraint can be solved by
\begin{align}
    \sum_{i=1}^{c}\lambda_{i}(L_{A})=Tr(F^{T}{L_{A}}F)
\end{align}
where $F\in \mathbb{R^{m \times c}}$. Thus, the final form of the optimization objective function can be written as
\begin{align}
\label{eq:16}
    &\mathop{\min}_{A,F} \left \| A-W_{ll} \right \|^{2}_{F}+ \gamma\left \| A \right \|^{2}_{F}+2\beta Tr(F^{T}{L_{A}}F)\\ \notag
& \begin{array}{r@{\quad}l@{}l@{\quad}l}
s.t.& \sum_{j}{A_{ij}=1}, A_{ij}\ge 0, F^{T}F=I. \\ 
\end{array}
\end{align}
In order to address this problem, the matrices $A$ and $F$ are updated in an alternating manner.

\textbf{Update $A$}: When $F$ is fixed, we have\cite{nie2016unsupervised}
\begin{align}
    Tr(F^{T}{L_{A}}F)=\frac{1}{2}\sum^{m}_{i,j=1} \left \|  f_{i}-f_{j} \right \| ^{2}_{2}a_{ij}
\end{align}
where $f_{i}$ and $f_{j}$ represent the $i_{th}$ and $j_{th}$ rows of $F$, respectively. Then we rewrite Eq.\ref{eq:16} as
\begin{align}
\label{eq:18}
    &\mathop{\min}_{A,F} \sum^{m}_{j=1}\left \| A_{i}-w^{ll}_{i} \right \|^{2}_{2}+ \gamma\sum^{m}_{j=1}A_{ij}^{2}+\beta\sum^{m}_{j=1} \left \|  f_{i}-f_{j} \right \| ^{2}_{2}a_{ij} \\ \notag
& \begin{array}{r@{\quad}l@{}l@{\quad}l}
s.t.& \sum_{j}{A_{ij}=1}, A_{ij}\ge 0, F^{T}F=I. \\ 
\end{array}
\end{align}
Given $e_{ij} = \beta\left \|  f_{i}-f_{j} \right \| ^{2}_{2}-2w_{ij}^{ll}$, Eq.\ref{eq:18} can be reformulated as
\begin{align}
\label{eq:19}
    \mathop{\min}_{ \sum_{j}{A_{ij}=1},A_{i}\ge 0} \frac{1}{2}\left \|  A_{i}+\frac{e_{i}}{2 (\gamma +1) } \right \| ^{2}_{2}
\end{align}
where $w^{ll}_{i}$ is the $i_{th}$ row of $W_{ll}$. The resolution of Eq. \ref{eq:19} can be addressed through the application of the Lagrange multiplier method
\begin{align}
\label{eq:20}
    A_{ij}=(-\frac{e_{ij}}{2(\gamma+1)}+\eta)_{+}
\end{align}
where $\eta$ is the Lagrange multiplier. We employ the closed-form solution presented in \cite{nie2016constrained} to formulate the optimal affinities $A_{ij}$ for Eq.\ref{eq:20}
\begin{align}
\label{eq:21}
    A_{ij} =
    \begin{cases}
        \frac{e_{i,h+1}-e_{ij}}{he_{i,h+1}-\sum^{h}_{n=1}e_{in}} &j\leq h\\
        0 &j> h
    \end{cases}
\end{align}
where $h$ is the number of non-zero values in each row of $A$.

\textbf{Update $F$}: When $A$ is fixed, Eq.\ref{eq:16} can be written as
\begin{align}
    \mathop{\min}_{F\in \mathbb{R}^{m\times c}, F^{T}F=I} Tr(F^{T}{L_{A}}F)
\end{align}
The optimal solution for $F$ is constituted by the $c$ eigenvectors of $L_{A}=D_{A}-\frac{A^{T}+A}{2}$ that correspond to the $c$ smallest eigenvalues.

After iterative updates, a similarity matrix $A$ is derived. The matrix encompasses $c$ connected components, which can be utilized for clustering directly. The complete sequence of the algorithm is delineated in Algorithm \ref{alg:2}. Upon obtaining the anchor label, the label information can be propagated to all unlabeled pixels following the procedure defined in Sections \ref{subsection:anchor graph} and \ref{subsection:KNN Graph}, thus achieving the classification of the hyperspectral image.

\begin{algorithm}
	\renewcommand{\algorithmicrequire}{\textbf{Input:}}
	\renewcommand{\algorithmicensure}{\textbf{Output:}}
        \caption{Optimization to Solve Eq.\ref{eq:13}.}
	\label{alg:2}
	\begin{algorithmic}[1]
            \REQUIRE The affinity matrix $W_{ll}\in \mathbb{R}^{m \times m}$, cluster number $c$, a large enough $\beta$, non-zero value number $k$ in each row.
            \ENSURE Similarity matrix $A\in \mathbb{R}^{m \times m}$ with exactly $k$ connected components.
            \STATE Initialize $F$, constituted by the $k$ eigenvectors of $L_{W_{ll}} = D_{W_{ll}}-W_{ll}$ that correspond to the $k$ smallest eigenvalues.     
            \WHILE{not converge}
                \STATE Update $\textbf{A}$ by Eq.\ref{eq:21};
                \STATE Update $\textbf{F}$ by perform eigenvalue decomposition on $L_{A}=D_{A}-(A^{T}+A)/2$;
            \ENDWHILE
	\end{algorithmic}  
\end{algorithm}

\subsection{Time Complexity Analysis}
To accelerate the computation and further diminish the time complexity, we employ Principal Component Analysis (PCA) to reduce the dimension of the original dataset to $d$ and partition the entire pixel set into several parts at a fixed interval $\theta$ after anchor selection. The first step of generating $m$ anchors by $k$-means requires a time complexity of $O(m(m+n)dt_{1})$, where $t_{1}$ is the number of iterations of the $k$-means. During anchor graph construction, it is essential to compute the similarity between anchor points and individual pixels. The computational time complexity for this procedure is $O({mnd})$. Given that $m$ is significantly smaller than $n$, it becomes evident that our initial stage of the anchor graph-based label propagation process exhibits linear-time complexity. This indicates that we are able to perform the preliminary classification of unlabeled data with high efficiency while maintaining reliable accuracy. During the second stage of label propagation, which is based on the derived sparse graph, the adoption of a closed-form solution eliminates the need for repetitive iterative processes, thus eliminating the associated computational loop. The primary computational expense of this phase lies in the construction of the affinity matrix $W$ for the entire set of samples. The matrix $W$ is decomposed into four sub-matrices, of which the computation is essential for sub-matrices $W_{ll}$ and $W_{uu}$. For $W_{ll}$, we only need the constant time complexity of $O(m^{2}d)$. For $W_{uu}$, currently, the construction of the sparse KNN graph requires the computation of similarities between all samples before identifying the $k-$nearest neighbors, which involves a time complexity of $O(n^{2}d)$ for the formation of the graph. Due to the implementation of the proposed top$k$-based pre-pruning strategy, the construction of $W_{uu}$ requires only a linear complexity $O(knd)$. The computational complexity associated with clustering anchor points is primarily dependent on the number of iterations and the computational costs to solve $A$ and $F$. Given that $m\ll n$, the approximation of the solutions for both $A$ and $F$ can be achieved with a constant time complexity of $O(1)$. Following $t_{2}$ iterations, the clustering algorithm demonstrates a total time complexity at the $O(t_{2})$ level. In conclusion, the classification algorithm we propose for the hyperspectral image onboard exhibits a linear-time complexity.

\section{Code availability}
The code implemented in this work will be available via GitHub.



\bibliography{ref}

@article{zhang2019label,
  title={Label propagation ensemble for hyperspectral image classification},
  author={Zhang, Youqiang and Cao, Guo and Shafique, Ayesha and Fu, Peng},
  journal={IEEE Journal of Selected Topics in Applied Earth Observations and Remote Sensing},
  volume={12},
  number={9},
  pages={3623--3636},
  year={2019},
  publisher={IEEE}
}

@article{jia2020superpixel,
  title={Superpixel-level weighted label propagation for hyperspectral image classification},
  author={Jia, Sen and Deng, Xianglong and Xu, Meng and Zhou, Jun and Jia, Xiuping},
  journal={IEEE Transactions on Geoscience and Remote Sensing},
  volume={58},
  number={7},
  pages={5077--5091},
  year={2020},
  publisher={IEEE}
}

@article{zhong2021dynamic,
  title={Dynamic spectral--spatial Poisson learning for hyperspectral image classification with extremely scarce labels},
  author={Zhong, Shengwei and Zhou, Tao and Wan, Sheng and Yang, Jian and Gong, Chen},
  journal={IEEE Transactions on Geoscience and Remote Sensing},
  volume={60},
  pages={1--15},
  year={2021},
  publisher={IEEE}
}

@article{wang2017fast,
  title={Fast spectral clustering with anchor graph for large hyperspectral images},
  author={Wang, Rong and Nie, Feiping and Yu, Weizhong},
  journal={IEEE Geoscience and Remote Sensing Letters},
  volume={14},
  number={11},
  pages={2003--2007},
  year={2017},
  publisher={IEEE}
}

@article{jiang2024structured,
  title={Structured Anchor Learning for Large-Scale Hyperspectral Image Projected Clustering},
  author={Jiang, Guozhu and Zhang, Yongshan and Wang, Xinxin and Jiang, Xinwei and Zhang, Lefei},
  journal={IEEE Transactions on Circuits and Systems for Video Technology},
  year={2024},
  publisher={IEEE}
}

@article{nie2023fast,
  title={Fast clustering with anchor guidance},
  author={Nie, Feiping and Xue, Jingjing and Yu, Weizhong and Li, Xuelong},
  journal={IEEE Transactions on Pattern Analysis and Machine Intelligence},
  volume={46},
  number={4},
  pages={1898--1912},
  year={2023},
  publisher={IEEE}
}

@inproceedings{liu2010large,
  title={Large graph construction for scalable semi-supervised learning},
  author={Liu, Wei and He, Junfeng and Chang, Shih-Fu},
  booktitle={Proceedings of the 27th international conference on machine learning (ICML-10)},
  pages={679--686},
  year={2010},
  organization={Citeseer}
}

@article{zhou2003learning,
  title={Learning with local and global consistency},
  author={Zhou, Dengyong and Bousquet, Olivier and Lal, Thomas and Weston, Jason and Sch{\"o}lkopf, Bernhard},
  journal={Advances in neural information processing systems},
  volume={16},
  year={2003}
}

@inproceedings{nie2016constrained,
  title={The constrained laplacian rank algorithm for graph-based clustering},
  author={Nie, Feiping and Wang, Xiaoqian and Jordan, Michael and Huang, Heng},
  booktitle={Proceedings of the AAAI conference on artificial intelligence},
  volume={30},
  number={1},
  year={2016}
}

@article{fan1949theorem,
  title={On a theorem of Weyl concerning eigenvalues of linear transformations I},
  author={Fan, Ky},
  journal={Proceedings of the National Academy of Sciences},
  volume={35},
  number={11},
  pages={652--655},
  year={1949}
}

@inproceedings{nie2016unsupervised,
  title={Unsupervised feature selection with structured graph optimization},
  author={Nie, Feiping and Zhu, Wei and Li, Xuelong},
  booktitle={Proceedings of the AAAI conference on artificial intelligence},
  volume={30},
  number={1},
  year={2016}
}

@article{mohar1991laplacian,
  title={The Laplacian spectrum of graphs},
  author={Mohar, Bojan and Alavi, Y and Chartrand, G and Oellermann, Ortrud},
  journal={Graph theory, combinatorics, and applications},
  volume={2},
  number={871-898},
  pages={12},
  year={1991},
  publisher={Wiley}
}

@article{li2016hyperspectral,
  title={Hyperspectral image classification using deep pixel-pair features},
  author={Li, Wei and Wu, Guodong and Zhang, Fan and Du, Qian},
  journal={IEEE Transactions on Geoscience and Remote Sensing},
  volume={55},
  number={2},
  pages={844--853},
  year={2016},
  publisher={IEEE}
}

@phdthesis{10.5555/1104523,
author = {Zhu, Xiaojin and Lafferty, John and Rosenfeld, Ronald},
title = {Semi-supervised learning with graphs},
year = {2005},
isbn = {0542190591},
publisher = {Carnegie Mellon University},
address = {USA},
school = {Carnegie Mellon University},
note = {AAI3179046}
}

@inproceedings{calder2020poisson,
  title={Poisson learning: Graph based semi-supervised learning at very low label rates},
  author={Calder, Jeff and Cook, Brendan and Thorpe, Matthew and Slepcev, Dejan},
  booktitle={International Conference on Machine Learning},
  pages={1306--1316},
  year={2020},
  organization={PMLR}
}

@article{zhong2020fusion,
  title={Fusion of spectral--spatial classifiers for hyperspectral image classification},
  author={Zhong, Shengwei and Chen, Shuhan and Chang, Chein-I and Zhang, Ye},
  journal={IEEE Transactions on Geoscience and Remote Sensing},
  volume={59},
  number={6},
  pages={5008--5027},
  year={2020},
  publisher={IEEE}
}

@article{zheng2019hyperspectral,
  title={Hyperspectral image classification with small training sample size using superpixel-guided training sample enlargement},
  author={Zheng, Chengyong and Wang, Ningning and Cui, Jing},
  journal={IEEE Transactions on Geoscience and Remote Sensing},
  volume={57},
  number={10},
  pages={7307--7316},
  year={2019},
  publisher={IEEE}
}

@article{wang2019scalable,
  title={Scalable graph-based clustering with nonnegative relaxation for large hyperspectral image},
  author={Wang, Rong and Nie, Feiping and Wang, Zhen and He, Fang and Li, Xuelong},
  journal={IEEE Transactions on Geoscience and Remote Sensing},
  volume={57},
  number={10},
  pages={7352--7364},
  year={2019},
  publisher={IEEE}
}

@article{rafiezadeh2020hierarchical,
  title={Hierarchical sparse subspace clustering (HESSC): An automatic approach for hyperspectral image analysis},
  author={Rafiezadeh Shahi, Kasra and Khodadadzadeh, Mahdi and Tusa, Laura and Ghamisi, Pedram and Tolosana-Delgado, Raimon and Gloaguen, Richard},
  journal={Remote Sensing},
  volume={12},
  number={15},
  pages={2421},
  year={2020},
  publisher={MDPI}
}

@article{cai2022superpixel,
  title={Superpixel contracted neighborhood contrastive subspace clustering network for hyperspectral images},
  author={Cai, Yaoming and Zhang, Zijia and Ghamisi, Pedram and Ding, Yao and Liu, Xiaobo and Cai, Zhihua and Gloaguen, Richard},
  journal={IEEE Transactions on Geoscience and Remote Sensing},
  volume={60},
  pages={1--13},
  year={2022},
  publisher={IEEE}
}

@article{zhao2021superpixel,
  title={Superpixel-level global and local similarity graph-based clustering for large hyperspectral images},
  author={Zhao, Haishi and Zhou, Fengfeng and Bruzzone, Lorenzo and Guan, Renchu and Yang, Chen},
  journal={IEEE Transactions on Geoscience and Remote Sensing},
  volume={60},
  pages={1--16},
  year={2021},
  publisher={IEEE}
}

@article{chen2023spectral,
  title={Spectral-spatial superpixel anchor graph-based clustering for hyperspectral imagery},
  author={Chen, Xiaohong and Zhang, Yongshan and Feng, Xuxiang and Jiang, Xinwei and Cai, Zhihua},
  journal={IEEE Geoscience and Remote Sensing Letters},
  volume={20},
  pages={1--5},
  year={2023},
  publisher={IEEE}
}

@article{paoletti2019deep,
  title={Deep learning classifiers for hyperspectral imaging: A review},
  author={Paoletti, Mercedes Eugenia and Haut, Juan Mario and Plaza, Javier and Plaza, Antonio},
  journal={ISPRS Journal of Photogrammetry and Remote Sensing},
  volume={158},
  pages={279--317},
  year={2019},
  publisher={Elsevier}
}

@article{nalepa2021towards,
  title={Towards on-board hyperspectral satellite image segmentation: Understanding robustness of deep learning through simulating acquisition conditions},
  author={Nalepa, Jakub and Myller, Michal and Cwiek, Marcin and Zak, Lukasz and Lakota, Tomasz and Tulczyjew, Lukasz and Kawulok, Michal},
  journal={Remote sensing},
  volume={13},
  number={8},
  pages={1532},
  year={2021},
  publisher={MDPI}
}

@inproceedings{aggarwal2015mixed,
  title={Mixed Gaussian and impulse denoising of hyperspectral images},
  author={Aggarwal, Hemant Kumar and Majumdar, Angshul},
  booktitle={2015 IEEE International Geoscience and Remote Sensing Symposium (IGARSS)},
  pages={429--432},
  year={2015},
  organization={IEEE}
}

@inproceedings{tariyal2015hyperspectral,
  title={Hyperspectral impulse denoising with sparse and low-rank penalties},
  author={Tariyal, Snigdha and Aggarwal, Hemant Kumar and Majumdar, Angshul},
  booktitle={2015 7th Workshop on Hyperspectral Image and Signal Processing: Evolution in Remote Sensing (WHISPERS)},
  pages={1--4},
  year={2015},
  organization={IEEE}
}

@article{rasti2018noise,
  title={Noise reduction in hyperspectral imagery: Overview and application},
  author={Rasti, Behnood and Scheunders, Paul and Ghamisi, Pedram and Licciardi, Giorgio and Chanussot, Jocelyn},
  journal={Remote Sensing},
  volume={10},
  number={3},
  pages={482},
  year={2018},
  publisher={MDPI}
}

@article{sellars2020superpixel,
  title={Superpixel contracted graph-based learning for hyperspectral image classification},
  author={Sellars, Philip and Aviles-Rivero, Angelica I and Sch{\"o}nlieb, Carola-Bibiane},
  journal={IEEE Transactions on Geoscience and Remote Sensing},
  volume={58},
  number={6},
  pages={4180--4193},
  year={2020},
  publisher={IEEE}
}

@inproceedings{evans2014ops,
  title={OPS-SAT: A ESA nanosatellite for accelerating innovation in satellite control},
  author={Evans, David and Merri, Mario},
  booktitle={SpaceOps 2014 Conference},
  pages={1702},
  year={2014}
}

@article{denby2019orbital,
  title={Orbital edge computing: Machine inference in space},
  author={Denby, Bradley and Lucia, Brandon},
  journal={IEEE Computer Architecture Letters},
  volume={18},
  number={1},
  pages={59--62},
  year={2019},
  publisher={IEEE}
}

@article{furano2020towards,
  title={Towards the use of artificial intelligence on the edge in space systems: Challenges and opportunities},
  author={Furano, Gianluca and Meoni, Gabriele and Dunne, Aubrey and Moloney, David and Ferlet-Cavrois, Veronique and Tavoularis, Antonis and Byrne, Jonathan and Buckley, L{\'e}onie and Psarakis, Mihalis and Voss, Kay-Obbe and others},
  journal={IEEE Aerospace and Electronic Systems Magazine},
  volume={35},
  number={12},
  pages={44--56},
  year={2020},
  publisher={IEEE}
}

@article{george2018onboard,
  title={Onboard processing with hybrid and reconfigurable computing on small satellites},
  author={George, Alan D and Wilson, Christopher M},
  journal={Proceedings of the IEEE},
  volume={106},
  number={3},
  pages={458--470},
  year={2018},
  publisher={IEEE}
}

@article{giuffrida2021varphi,
  title={The $\Phi$-Sat-1 mission: The first on-board deep neural network demonstrator for satellite earth observation},
  author={Giuffrida, Gianluca and Fanucci, Luca and Meoni, Gabriele and Bati{\v{c}}, Matej and Buckley, L{\'e}onie and Dunne, Aubrey and Van Dijk, Chris and Esposito, Marco and Hefele, John and Vercruyssen, Nathan and others},
  journal={IEEE Transactions on Geoscience and Remote Sensing},
  volume={60},
  pages={1--14},
  year={2021},
  publisher={IEEE}
}

@article{justo2024semantic,
  title={Semantic Segmentation in Satellite Hyperspectral Imagery by Deep Learning},
  author={Justo, Jon Alvarez and Ghita, Alexandru and Kovac, Daniel and Garrett, Joseph L and Georgescu, Mariana-Iuliana and Gonzalez-Llorente, Jesus and Ionescu, Radu Tudor and Johansen, Tor Arne},
  journal={IEEE Journal of Selected Topics in Applied Earth Observations and Remote Sensing},
  year={2024},
  publisher={IEEE}
}

@article{langer2023robust,
  title={Robust and reconfigurable on-board processing for a hyperspectral imaging small satellite},
  author={Langer, Dennis D and Orlandi{\'c}, Milica and Bakken, Sivert and Birkeland, Roger and Garrett, Joseph L and Johansen, Tor A and S{\o}rensen, Asgeir J},
  journal={Remote Sensing},
  volume={15},
  number={15},
  pages={3756},
  year={2023},
  publisher={MDPI}
}

@article{morcillo2024parametric,
  title={Parametric pipelined k-means implementation for hyperspectral processing on spacecraft embedded fpga},
  author={Morcillo, Borja and B{\'a}scones, Daniel and Gonz{\'a}lez, Carlos and Mend{\'\i}as, Jos{\'e} M and Mozos, Daniel},
  journal={IEEE Journal of Selected Topics in Applied Earth Observations and Remote Sensing},
  year={2024},
  publisher={IEEE}
}

@inproceedings{kieffer1996detection,
  title={Detection and correction of bad pixels in hyperspectral sensors},
  author={Kieffer, Hugh H},
  booktitle={Hyperspectral Remote Sensing and Applications},
  volume={2821},
  pages={93--108},
  year={1996},
  organization={SPIE}
}

@article{santos2022geometric,
  title={Geometric calibration of a hyperspectral frame camera with simultaneous determination of sensors misalignment},
  author={Santos, Lucas D and Tommaselli, Antonio MG and Berveglieri, Adilson and Imai, Nilton N and Oliveira, Raquel A and Honkavaara, Eija},
  journal={ISPRS Open Journal of Photogrammetry and Remote Sensing},
  volume={4},
  pages={100015},
  year={2022},
  publisher={Elsevier}
}

@article{zhuang2021fasthymix,
  title={FastHyMix: Fast and parameter-free hyperspectral image mixed noise removal},
  author={Zhuang, Lina and Ng, Michael K},
  journal={IEEE Transactions on Neural Networks and Learning Systems},
  volume={34},
  number={8},
  pages={4702--4716},
  year={2021},
  publisher={IEEE}
}

@article{ma2024spatial,
  title={Spatial pooling transformer network and noise-tolerant learning for noisy hyperspectral image classification},
  author={Ma, Jingjing and Zou, Yizhou and Tang, Xu and Zhang, Xiangrong and Liu, Fang and Jiao, Licheng},
  journal={IEEE Transactions on Geoscience and Remote Sensing},
  year={2024},
  publisher={IEEE}
}

@article{zhu2023collaborative,
  title={Collaborative hyperspectral image processing using satellite edge computing},
  author={Zhu, Botao and Lin, Siyuan and Zhu, Yifei and Wang, Xudong},
  journal={IEEE Transactions on Mobile Computing},
  volume={23},
  number={3},
  pages={2241--2253},
  year={2023},
  publisher={IEEE}
}

@inproceedings{kruse2014multispectral,
  title={Multispectral, hyperspectral, and LiDAR remote sensing and geographic information fusion for improved earthquake response},
  author={Kruse, FA and Kim, AM and Runyon, SC and Carlisle, Sarah C and Clasen, CC and Esterline, CH and Jalobeanu, Andr{\'e} and Metcalf, JP and Basgall, PL and Trask, DM and others},
  booktitle={Algorithms and Technologies for Multispectral, Hyperspectral, and Ultraspectral Imagery XX},
  volume={9088},
  pages={132--145},
  year={2014},
  organization={SPIE}
}

@article{arias2019hyperspectral,
  title={Hyperspectral imaging retrieval using MODIS satellite sensors applied to volcanic ash clouds monitoring},
  author={Arias, Luis and Cifuentes, Jose and Mar{\'\i}n, Milton and Castillo, Fernando and Garc{\'e}s, Hugo},
  journal={Remote sensing},
  volume={11},
  number={11},
  pages={1393},
  year={2019},
  publisher={MDPI}
}

@article{zhang2024uav,
  title={UAV Hyperspectral Remote Sensing Image Classification: A Systematic Review},
  author={Zhang, Zhen and Huang, Lehao and Wang, Qingwang and Jiang, Linhuan and Qi, Yemao and Wang, Shunyuan and Shen, Tao and Tang, Bo-Hui and Gu, Yanfeng},
  journal={IEEE Journal of Selected Topics in Applied Earth Observations and Remote Sensing},
  year={2024},
  publisher={IEEE}
}

@article{audebert2019deep,
  title={Deep learning for classification of hyperspectral data: A comparative review},
  author={Audebert, Nicolas and Le Saux, Bertrand and Lef{\`e}vre, S{\'e}bastien},
  journal={IEEE geoscience and remote sensing magazine},
  volume={7},
  number={2},
  pages={159--173},
  year={2019},
  publisher={IEEE}
}

@article{chen2011simple,
  title={A simple and effective method for filling gaps in Landsat ETM+ SLC-off images},
  author={Chen, Jin and Zhu, Xiaolin and Vogelmann, James E and Gao, Feng and Jin, Suming},
  journal={Remote sensing of environment},
  volume={115},
  number={4},
  pages={1053--1064},
  year={2011},
  publisher={Elsevier}
}

@article{wang2021filling,
  title={Filling gaps in Landsat ETM+ SLC-off images with Sentinel-2 MSI images},
  author={Wang, Qunming and Wang, Lanxing and Wei, Chao and Jin, Yanmin and Li, Zhongbin and Tong, Xiaohua and Atkinson, Peter M},
  journal={International Journal of Applied Earth Observation and Geoinformation},
  volume={101},
  pages={102365},
  year={2021},
  publisher={Elsevier}
}

@article{zhang2019hybrid,
  title={Hybrid noise removal in hyperspectral imagery with a spatial--spectral gradient network},
  author={Zhang, Qiang and Yuan, Qiangqiang and Li, Jie and Liu, Xinxin and Shen, Huanfeng and Zhang, Liangpei},
  journal={IEEE Transactions on Geoscience and Remote Sensing},
  volume={57},
  number={10},
  pages={7317--7329},
  year={2019},
  publisher={IEEE}
}

@article{aggarwal2016hyperspectral,
  title={Hyperspectral image denoising using spatio-spectral total variation},
  author={Aggarwal, Hemant Kumar and Majumdar, Angshul},
  journal={IEEE Geoscience and Remote Sensing Letters},
  volume={13},
  number={3},
  pages={442--446},
  year={2016},
  publisher={IEEE}
}

@article{fang2025pcet,
  title={PCET: Patch Confidence-Enhanced Transformer with efficient spectral--spatial features for hyperspectral image classification},
  author={Fang, Li and Lan, Xuanli and Li, Tianyu and Shen, Huifang},
  journal={International Journal of Applied Earth Observation and Geoinformation},
  volume={136},
  pages={104308},
  year={2025},
  publisher={Elsevier}
}

@article{shi2025few,
  title={Few-Shot Learning Based on Multilevel Contrast for Cross-domain Hyperspectral Image Classification},
  author={Shi, Cheng and Liu, Weijun and Fang, Li and You, Zhenzhen and Miao, Qiguang and Pun, Chi-Man},
  journal={IEEE Transactions on Geoscience and Remote Sensing},
  year={2025},
  publisher={IEEE}
}

@article{shi2025multil,
  title={Multil-scale Spatial-Frequency Domain Cross Transformer for Hyperspectral Image Classification},
  author={Shi, Cheng and Chen, Pupu and Fang, Li and Zhao, Minghua and Hei, Xinhong and Miao, Qiguang},
  journal={IEEE Transactions on Instrumentation and Measurement},
  year={2025},
  publisher={IEEE}
}

@article{shi2022explainable,
  title={Explainable scale distillation for hyperspectral image classification},
  author={Shi, Cheng and Fang, Li and Lv, Zhiyong and Zhao, Minghua},
  journal={Pattern Recognition},
  volume={122},
  pages={108316},
  year={2022},
  publisher={Elsevier}
}

@article{he2024ub,
  title={UB-FineNet: Urban building fine-grained classification network for open-access satellite images},
  author={He, Zhiyi and Yao, Wei and Shao, Jie and Wang, Puzuo},
  journal={ISPRS Journal of Photogrammetry and Remote Sensing},
  volume={217},
  pages={76--90},
  year={2024},
  publisher={Elsevier}
}

@article{jiang2023space,
  title={Space-to-speed architecture supporting acceleration on VHR image processing},
  author={Jiang, Shenlu and Tarabalka, Yuliya and Yao, Wei and Hong, Zhonghua and Feng, Guofu},
  journal={ISPRS Journal of Photogrammetry and Remote Sensing},
  volume={198},
  pages={30--44},
  year={2023},
  publisher={Elsevier}
}

@article{sun2020deep,
  title={Deep clustering with intraclass distance constraint for hyperspectral images},
  author={Sun, Jinguang and Wang, Wanli and Wei, Xian and Fang, Li and Tang, Xiaoliang and Xu, Yusheng and Yu, Hui and Yao, Wei},
  journal={IEEE Transactions on Geoscience and Remote Sensing},
  volume={59},
  number={5},
  pages={4135--4149},
  year={2020},
  publisher={IEEE}
}

@article{zhang20191d,
  title={1D-convolutional capsule network for hyperspectral image classification},
  author={Zhang, Haitao and Meng, Lingguo and Wei, Xian and Tang, Xiaoliang and Tang, Xuan and Wang, Xingping and Jin, Bo and Yao, Wei},
  journal={arXiv preprint arXiv:1903.09834},
  year={2019}
}
\end{document}